\documentclass[10pt,journal,compsoc]{IEEEtran}

\ifCLASSOPTIONcompsoc
  \usepackage[nocompress]{cite}
\else
  \usepackage{cite}
\fi
\ifCLASSINFOpdf
\else
\fi
\usepackage{graphicx}
\usepackage{xcolor}
\usepackage{color}
\usepackage{mathrsfs}
\usepackage{amsmath}
\usepackage{amssymb}
\usepackage{ragged2e}
\usepackage{algorithm}
\usepackage{listings}

\def\ie{\emph{i.e.}}
\def\eg{\emph{e.g.}}

\def\etc{{\em etc.}} 
\definecolor{dkgreen}{rgb}{0,0.6,0}
\definecolor{gray}{rgb}{0.5,0.5,0.5}
\definecolor{mauve}{rgb}{0.58,0,0.82}
\newcommand{\para}[1]{\vspace{.05in}\noindent\textbf{#1}}

\usepackage{booktabs}
\usepackage{array, caption, threeparttable}
\usepackage[font=small,labelfont=bf]{caption}
\begin{document}
\title{MCIBI++: Soft Mining Contextual Information Beyond Image for Semantic Segmentation}
\author{Zhenchao Jin,
        Dongdong Yu,
        Zehuan Yuan,
        and Lequan Yu}

\markboth{Journal of \LaTeX\ Class Files,~Vol.~14, No.~8, August~2015}
{Shell \MakeLowercase{\textit{et al.}}: Bare Demo of IEEEtran.cls for Computer Society Journals}
\IEEEtitleabstractindextext{
\begin{abstract}
    \justifying
    Co-occurrent visual pattern makes context aggregation become an essential paradigm for semantic segmentation.
    The existing studies focus on modeling the contexts within image while neglecting the valuable semantics of the corresponding category beyond image.
    To this end, we propose a novel soft mining contextual information beyond image paradigm named MCIBI++ to further boost the pixel-level representations. 
    Specifically, we first set up a dynamically updated memory module to store the dataset-level distribution information of various categories and then leverage the information to yield the dataset-level category representations during network forward.
    After that, we generate a class probability distribution for each pixel representation and conduct the dataset-level context aggregation with the class probability distribution as weights.
    Finally, the original pixel representations are augmented with the aggregated dataset-level and the conventional image-level contextual information.
    Moreover, in the inference phase, we additionally design a coarse-to-fine iterative inference strategy to further boost the segmentation results.
    MCIBI++ can be effortlessly incorporated into the existing segmentation frameworks and bring consistent performance improvements.
    Also, MCIBI++ can be extended into the video semantic segmentation framework with considerable improvements over the baseline.
    Equipped with MCIBI++, we achieved the state-of-the-art performance on seven challenging image or video semantic segmentation benchmarks.
\end{abstract}

\begin{IEEEkeywords}
   Semantic Segmentation, Context Aggregation, Dataset Statistics, Memory Module, Iterative Testing, Video Semantic Segmentation
\end{IEEEkeywords}}

\maketitle

\IEEEdisplaynontitleabstractindextext

\IEEEpeerreviewmaketitle

\IEEEraisesectionheading{\section{Introduction}\label{sec:introduction}}

\IEEEPARstart{S}{emantic} segmentation is a long-standing challenging task in computer vision field, aiming to assign a semantic label to each pixel in an image.
This task is of broad interest for potential applications of autonomous driving, medical diagnosing, robot sensing, to name a few.
In recent years, deep neural networks~\cite{simonyan2014very,he2016deep} have become the dominant solutions \cite{cao2019gcnet,zhao2017pyramid,chen2017deeplab,chen2017rethinking,chen2018encoder,fu2019dual,wang2018non,berman2018lovasz},
where the encoder-decoder architecture~\cite{long2015fully} is the cornerstone of these methods.
Specifically, the current studies include adopting the graphical models or the cascade structure to refine the segmentation results~\cite{cheng2020cascadepsp,chen2014semantic,zheng2015conditional,kirillov2020pointrend,ruan2019devil}, 
designing novel network backbones to extract more discriminative feature representations~\cite{yu2017dilated,zhang2020resnest,wang2020deep,liu2021swin,zheng2021rethinking},
replacing the conventional CNN-based decoder with the transformer-based decoder \cite{cheng2021per},
aggregating the contextual information to augment the original pixel-level representations~\cite{chen2017deeplab,zhao2017pyramid,wang2018non,jin2021isnet,fu2019adaptive,yu2020context}, and so on.

As there exist co-occurrent visual patterns in images, modeling the context, \ie, relations of different pixels, is universal and essential for semantic segmentation.
Prior to this work, the context of a pixel typically refers to a set of other pixels in the input image, \eg, the surrounding pixels.
Specifically, PSPNet~\cite{zhao2017pyramid} utilizes the pyramid spatial pooling to aggregate the contextual information within the input image.
DeepLab~\cite{chen2017deeplab,chen2017rethinking,chen2018encoder} exploits the contextual information by introducing the atrous spatial pyramid pooling (ASPP).
OCRNet~\cite{yuan2019object} proposes to improve the representation of each pixel by the weighted aggregation of the object region representations.
Several recent works~\cite{fu2019dual,yuan2018ocnet,huang2019ccnet,wang2018non} calculate the relations between a pixel and its contextual pixels and then aggregate the representations of the contextual pixels with higher weights for similar representations.
Moreover, some studies~\cite{berman2018lovasz,ke2018adaptive,zhao2019region} find that the pixel-wise cross entropy loss fundamentally lacks the spatial discrimination power and thereby propose to verify the segmentation structures directly with novel context-aware optimization objectives.
Nonetheless, all of the existing approaches only model the context within individual images, while neglecting the valuable contextual information beyond the input image.
The deep neural network-based semantic segmentation framework essentially categorizes the representations of pixels from the perspective of \textit{the whole dataset} rather than \textit{the single input image} in a non-linear embedding space.
Therefore, to categorize a pixel accurately, the semantic information of the corresponding category in the other images should also be considered.

\begin{figure}
\centering
\includegraphics[width=0.48\textwidth]{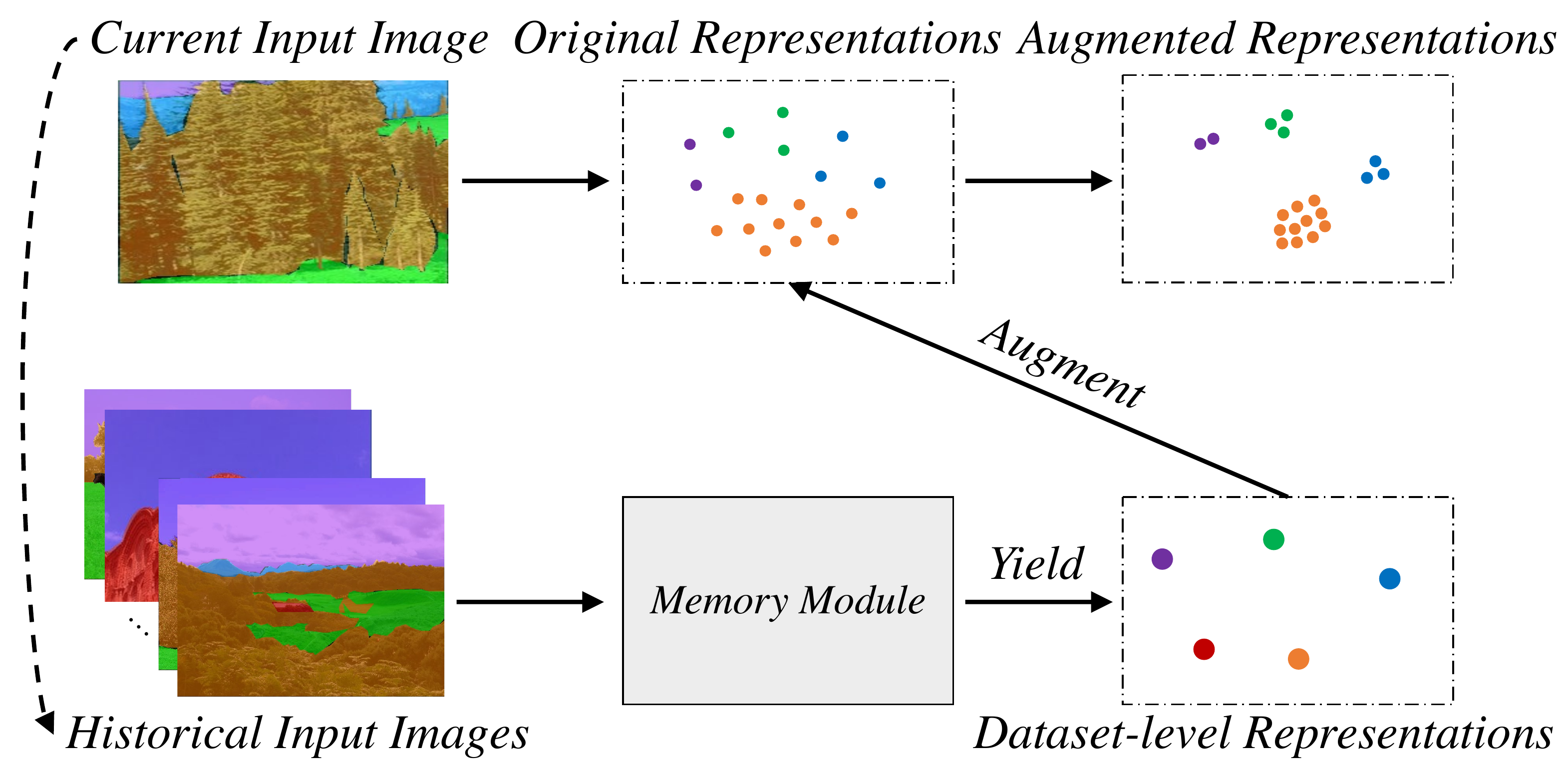}
\vspace{-0.3cm}
\caption{
   Illustration of soft mining contextual information beyond the input image (MCIBI++).
   The \textbf{memory module} stores the distribution information of the dataset-level representations of various categories.
   The dotted line shows that the \textbf{current input image} will be added into the \textbf{historical input images} after current iteration during learning.
}\label{motivation}
\vspace{-0.45cm}
\end{figure}

To overcome the aforementioned limitation of existing approaches, this paper proposes to mine the contextual information beyond the input image for semantic segmentation so that the pixel representations can be further improved.
As illustrated in Figure~\ref{motivation}, we first set up a memory module to store the dataset-level distribution information of various categories.
Inspired by the network normalization techniques \cite{ba2016layer,ioffe2015batch,ulyanov2016instance,wu2018group}, the distribution information in the memory module is dynamically updated according to the input images during the training while fixed in the inference phase.
With the learned distribution information in the memory module, we further generate the dataset-level category representations to boost the original pixel representations.
Specifically, we predict a class probability distribution for each pixel within the input image, where the distribution is learned under the supervision of the ground-truth segmentation.
After that, we conduct the dataset-level context aggregation with the weighted aggregation of the yielded dataset-level category representations, where the weights are determined by the generated class probability distribution.
Since the generated dataset-level representation contains the global statistic information of the corresponding category, it can incorporate more well-structured semantic information into the current input image to boost the segmentation performance from the perspective of the whole dataset.

\begin{table}[t]
\centering
\caption{
   Illustration of the motivation of proposing the coarse-to-fine iterative inference strategy.
   The segmentors are trained on ADE20K \emph{train split} and evaluated on \emph{validation set}.
   GT-MCIBI++ means using the ground-truth segmentation to conduct dataset-level context aggregation.
}\label{upperbound}
\vspace{-0.3cm}
\resizebox{.48\textwidth}{!}{
\begin{tabular}{c|c|c|c}
   \hline
   \hline
   Method                         &Backbone      &Stride        &mIoU ($\%$)       \\
   \hline         
   FCN                            &ResNet-50     &$8\times$     &36.96             \\
   FCN+MCIBI++                    &ResNet-50     &$8\times$     &43.39             \\
   FCN+GT-MCIBI++                 &ResNet-50     &$8\times$     &68.82             \\
   \hline
   \hline
\end{tabular}}
\vspace{-0.30cm}
\end{table}

Different from directly storing the dataset-level category representations, we propose to store the dataset-level distribution information of each category (\ie, mean and variances of category features) and utilize it to yield the demanded feature representations to reduce the memory usage during network training and boost the effectiveness of the dataset-level semantics.
Thus, we refer our proposed paradigm as soft mining contextual information beyond image (MCIBI++).
In addition, in our preliminary experiments as demonstrated in Table \ref{upperbound}, 
we observe that employing the ground-truth segmentation to conduct dataset-level context aggregation (FCN+GT-MCIBI++) outperforms that with predicted class probability distribution (FCN+MCIBI++) with a large margin (\ie, $25.43\%$ mIoU).
This result indicates that the predicted class probability distribution of the pixel representations may be insufficient to effectively aggregrate the dataset-level category representation.
We thereby adopt a coarse-to-fine iterative inference strategy to progressively improve the accuracy of aggregation weights so as to incorporate more effective dataset-level semantics for each pixel.

Our MCIBI++ paradigm can be seamlessly incorporated into the existing segmentation frameworks (\eg, FCN \cite{long2015fully}, ISNet \cite{jin2021isnet}, PSPNet \cite{zhao2017pyramid}, UperNet \cite{xiao2018unified} and DeepLabV3 \cite{chen2017rethinking}) and brings consistent yet solid performance improvements.
Equipped with MCIBI++, our whole framework achieves the state-of-the-art intersection-over-union segmentation scores (mIoU) performance on six challenging benchmarks, \ie, ADE20K \cite{zhou2017scene}, LIP \cite{gong2017look}, Cityscapes \cite{cordts2016cityscapes}, COCO-Stuff \cite{caesar2018coco}, PASCAL-Context \cite{everingham2010pascal} and PASCAL VOC 2012 \cite{everingham2015pascal, hariharan2011semantic}.
Furthermore, MCIBI++ can also be extended to the video semantic segmentation task, which needs to adopt the additional temporal information to obtain higher predictive accuracy.
Extensive experiments on the VSPW \cite{miao2021vspw} benchmark further demonstrate the effectiveness of our proposed paradigm.
We expect this work to inspire more researches on dataset-level context aggregation for semantic segmentation.

In a nutshell, our main contributions are summarized as:
\begin{itemize}
   \item This is one of the earliest works to explore how to mine the contextual information beyond the input image to further boost the segmentation performance. 
   Specifically, we propose MCIBI++ paradigm to aggregate the dataset-level representations of various categories to boost the pixel-level representations of the current image.
   \item We design a simple yet effective memory module to store the dataset-level distribution information of various categories and thus yield more discriminative dataset-level category representation.
   \item We propose a novel dataset-level context aggregation scheme to incorporate the dataset-level semantic information to boost the discriminative capability of the original pixel-level representations.
   \item We propose an effective coarse-to-fine iterative inference strategy to progressively incorporate more accurate dataset-level category representations for each pixel.
   \item With the proposed MCIBI++ paradigm, we report the state-of-the-art performance on six image semantic segmentation benchmarks and one video semantic segmentation benchmark.
\end{itemize}

\noindent \textbf{Difference from Conference Paper.}
A preliminary version of this manuscript was previously published in \cite{jin2021mining}.
In this expanded version, we significantly improve the conference version: 
(i) the design of the memory module, we replace the dataset-level category representations in the memory module with the dataset-level category distribution information and then employ the stored statistic information to yield the dataset-level category representations to reduce the memory size and improve the segmentation performance;
(ii) the strategy in the inference phase, we further design a coarse-to-fine iterative inference strategy to progressively boost the segmentation performance through incorporating the more and more accurate dataset-level category representations for the pixels.
(iii) we conduct additional experiments and analysis on ADE20K and Cityscapes datasets and compared with more recent methods to further show the state-of-the-art performance on six challenging benchmarks (\ie, ADE20K, LIP, Cityscapes, COCO-Stuff, PASCAL-Context and PASCAL VOC 2012);
(iv) we extend MCIBI++ to the video semantic segmentation task and also achieve consistent performance improvements and report the state-of-the-art results on the VSPW dataset.
Due to the merits of mining contextual information beyond the input image as well as the generality and flexibility of our method, 
we obtain the second place in The 1st Video Scene Parsing in the Wild Challenge Workshop, ICCV2021 \cite{miao2021vspw, jin2021memory} (the participants who won the first place used the extra training data from MS COCO \cite{lin2014microsoft});
(v) from both qualitative and quantitative perspective, our results are significantly better compared to our conference version \cite{jin2021mining}.
\section{Related Work}

\para{Semantic Segmentation.}
The deep neural networks~\cite{simonyan2014very,he2016deep} based methods have achieved remarkable success in semantic segmentation since the seminal work of fully convolutional networks (FCN) ~\cite{long2015fully}.
Modern semantic segmentation methods~\cite{chen2017deeplab,chen2017rethinking,chen2018encoder,zhao2017pyramid,cao2019gcnet,yuan2019object} focus on further improving the framework of FCN.
In details, to refine the coarse predictions of FCN, some researchers proposed to adopt the graphical models, such as CRF~\cite{liu2017deep,chen2014semantic,zheng2015conditional} and region growing~\cite{dias2018semantic}.
CascadePSP~\cite{cheng2020cascadepsp} proposed to refine the segmentation results with a general cascade segmentation refinement model.
Some novel backbone networks have also been designed to extract more effective feature representations~\cite{yu2017dilated,zhang2020resnest,wang2020deep}.
Feature pyramid methods \cite{kirillov2019panoptic,zhao2017pyramid,liu2015parsenet,chen2018encoder}, 
image pyramid methods \cite{chen2016attention,lin2017refinenet,lin2016efficient}, 
dilated convolutions \cite{chen2017deeplab,chen2017rethinking,chen2018encoder} 
and context-aware optimization \cite{berman2018lovasz,ke2018adaptive,zhao2019region}
are four popular paradigms to alleviate the limited receptive field issue or incorporate contextual information, in FCN.
More recently, the academic attention has been geared toward transformer \cite{vaswani2017attention} and 
many efforts \cite{beit,zheng2021rethinking,liu2021swin,xie2021segformer,strudel2021segmenter,cheng2021per} 
have been made to introduce the transformer architectures into image semantic segmentation.
Specifically, SETR \cite{zheng2021rethinking} proposed to improve the encoder-decoder based FCN architecture by encoding the input image as a sequence of patches.
Swin Transformer \cite{liu2021swin} and SegFormer \cite{xie2021segformer} proposed to construct the hierarchical features maps to enhance the pixel representations.
MaskFormer \cite{cheng2021per} reformulated the semantic-level and instance-level segmentation as a mask classification task and 
achieved excellent results in both semantic and panoptic segmentation tasks.
This work focuses on how to incorporate contextual information beyond image and the proposed paradigm can be utilized in both CNN-based and transformer-based segmentation approaches.

\para{Context Scheme.}
Co-occurrent visual pattern makes aggregating long-range context in the final feature map become a common practice in semantic segmentation networks.
To obtain the multi-scale contextual information, PSPNet~\cite{zhao2017pyramid} performs the spatial pooling at several grid scales, 
while DeepLab~\cite{chen2017deeplab,chen2017rethinking,chen2018encoder} proposed to adopt the pyramid dilated convolutions with different dilation rates.
Based on DeepLab, DenseASPP~\cite{yang2018denseaspp} further densified the dilated rates to cover larger scale ranges.
Recently, inspired by the self-attention scheme~\cite{vaswani2017attention}, 
many studies~\cite{zhao2018psanet,fu2019dual,yuan2018ocnet,cao2019gcnet,wang2018non,zhu2019asymmetric} proposed to augment the pixel representations by a weighted aggregation of all pixel representations in the input image, 
where the weights are determined by the similarities between the pixel representations.
Apart from this, some researches~\cite{yuan2019object,zhang2019acfnet,li2019expectation} proposed to first group the pixels into a set of regions and then 
improve the pixel representations by a weighted aggregation of these region representations, where the weights are the context relations between the pixel representations and the region representations.
Despite impressive, all the existing approaches only exploit the contextual information within individual images.
Different from these previous works, this paper proposes to mine more valuable contextual information for each pixel from the whole dataset to further improve the pixel representations.
Concurrently, there are some very recent studies~\cite{wang2021exploring,zhang2021looking,hu2021region} that share the similar motivation with our work.
Different from our method that aggregates the dataset-level contextual information beyond the input image via a memory module, these approaches extend the conventional contrastive learning strategy to the image semantic segmentation task to learn intra-class compact and inter-class dispersed features from the perspective of the whole dataset.

\para{Memory Module.}
Memory module enables the deep neural networks to capture effective information beyond the current input image.
It has shown power in various computer vision tasks~\cite{wang2020cross,chen2020memory,zhong2019invariance,li2019memory,oh2019video,wu2019long}.
For instance, MEGA~\cite{chen2020memory} adopted the memory mechanism to augment the candidate box features of key frame for video object detection, 
while MNE~\cite{li2019memory} leveraged the memory-based neighbourhood embedding to enhance the general convolutional features for image search and few shot learning tasks.
For deep metric learning, XBM~\cite{wang2020cross} designed a cross-batch memory module to mine informative examples across multiple mini-batches.
Distinct from these previous methods, the memory module in this paper is used to store the dataset-level distribution information (\ie, mean and variance) of various categories rather than the feature representations themselves and the demanded dataset-level category representations will be generated according to the corresponding distribution information.
In this case, we can capture the contextual information beyond the input image in a softer way to obtain more discriminative pixel-level representations from the perspective of the whole dataset.
To the best of our knowledge, we are the first to utilize the idea of the memory module in supervised image semantic segmentation.

\begin{figure*}
\centering
\includegraphics[width=0.96\textwidth]{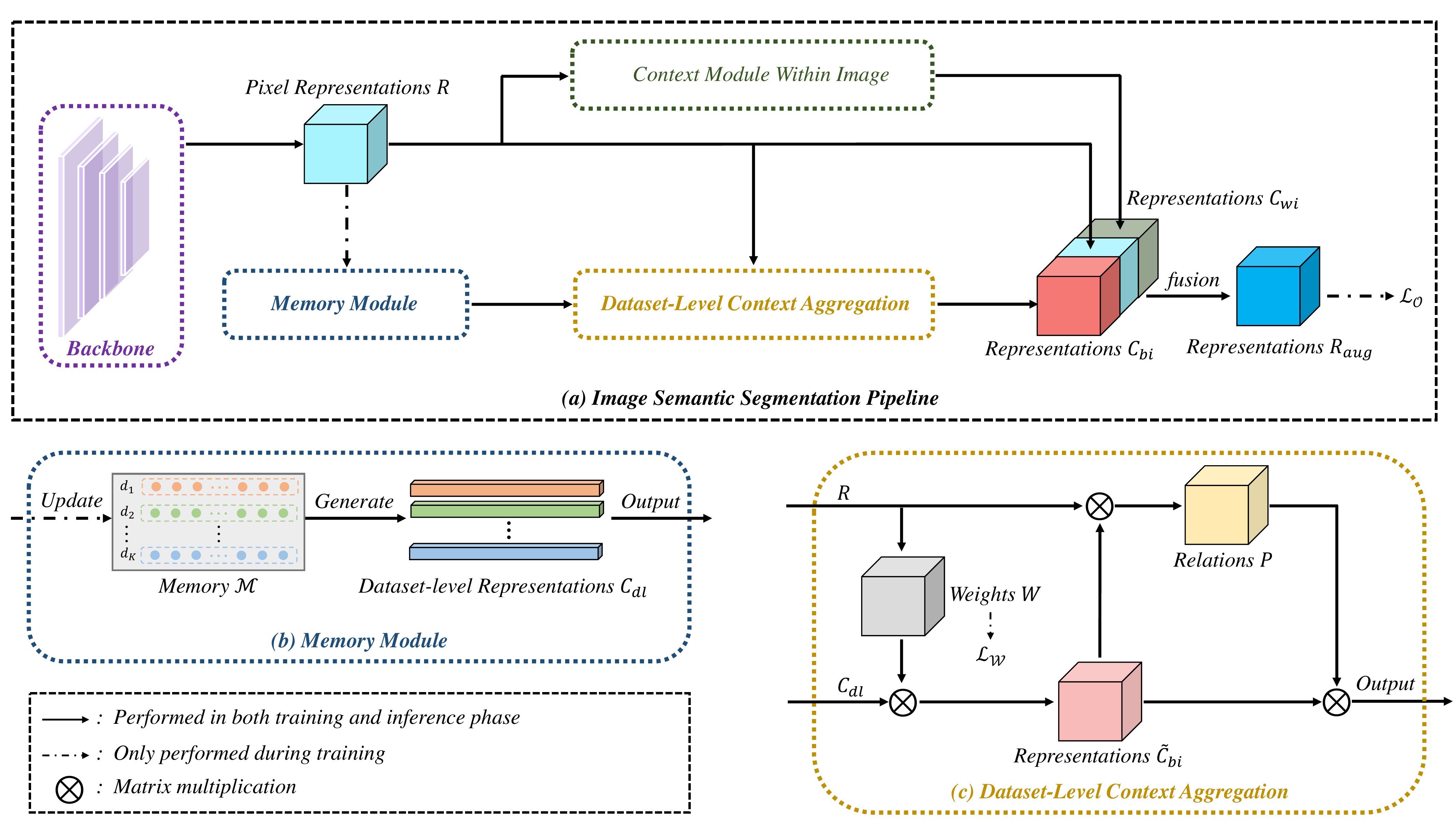}
\vspace{-0.30cm}
\caption{
    Illustrating the pipeline of soft mining contextual information beyond the input image.
    The \emph{Context Module Within Image} denotes for applying existing context scheme within the input image (\emph{e.g.}, PPM \cite{zhao2017pyramid}, ASPP \cite{chen2017deeplab} and UperNet \cite{xiao2018unified}), which is an optional operation.
    The \emph{Memory Module} is used to obtain the dataset-level distribution information of various categories and generate the dataset-level category representations.
    The \emph{Dataset-Level Context Aggregation} is designed to incorporate the yielded dataset-level category representations into the original pixel representations.
}\label{framework}
\vspace{-0.40cm}
\end{figure*}

\para{Video Semantic Segmentation.}
Video semantic segmentation aims to predict all pixels in a video with different semantic classes~\cite{miao2021vspw}.
Previous methods~\cite{gadde2017semantic,jain2019accel,li2018low,hu2020temporally,nilsson2018semantic,zhu2019improving,liu2017surveillance,miao2021vspw} can be divided into two streams. 
One stream is to explore the temporal relations of different frames via a warping module \cite{gadde2017semantic,zhu2017deep} so that the segmentation accuracy can be improved.
For example, DFF \cite{zhu2017deep} proposed to adopt a flow field to propagate the deep feature maps from key frames to other frames.
The other stream pays attention to improving the efficiency of the video semantic segmentation \cite{liu2020efficient,hu2020temporally,jain2019accel,li2018low}.
For instance, ETC \cite{liu2020efficient} proposed to leverage the knowledge distillation scheme to reduce the computing cost 
and Accel \cite{jain2019accel} proposed to combine the predictions of a reference branch to improve the segmentation efficiency.
Recently, some researchers \cite{miao2021vspw} introduced VSPW dataset, which is a large-scale video semantic segmentation benchmark.
Based on the benchmark dataset, they proposed a generic temporal context blending scheme and achieved solid results on VSPW by extending OCRNet \cite{yuan2019object} and PSPNet \cite{zhao2017pyramid} to the temporal dimension.
In this paper, we aim to boost the segmentation accuracy by extending MCIBI++ into the video semantic segmentation task.
Different from aggregating the pixel representations of some previous frames as the video context information~\cite{gadde2017semantic,hu2020temporally,miao2021vspw}, 
we exploit the temporal relations by aggregating the dataset-level representations of some previous frames to enhance the segmentation of the current one.
\section{Methodology}
In this section, we first introduce the paradigm of our soft mining contextual information beyond image (MCIBI++).
Then, we describe the details of the proposed coarse-to-fine iterative inference strategy.
Finally, we elaborate how to extend MCIBI++ to the video semantic segmentation task by adding the temporal information in the process of gathering the dataset-level representations for the pixels.

\subsection{Overview of MCIBI++} 
Given an input RGB image $I \in \mathbb{R}^{3 \times H \times W}$, we first project the pixels in $I$ into a non-linear embedding space with a backbone network $\mathcal{F}_{B}$ as follows:
\begin{equation} \label{eq1}
    R = \mathcal{F}_{B}(I),
\end{equation}
where the matrix $R$ with the size $Z \times \frac{H}{8} \times \frac{W}{8}$ denotes the pixel representations of $I$ and the feature dimension of the pixel representation is $Z$.
As shown in Fig.~\ref{framework}, to mine the contextual information beyond the input image $C_{bi}$, we conduct the dataset-level context aggregation:
\begin{equation} \label{eq2}
   C_{bi} = \mathcal{A}_{bi}(\mathcal{H}_{pr}(R), ~\mathcal{G}(\mathcal{M})),
\end{equation}
where $\mathcal{G}$ is utilized to yield the dataset-level representations of various categories according to the dataset-level distribution information stored in the memory module $\mathcal{M}$.
$\mathcal{A}_{bi}$ is the proposed dataset-level context aggregation scheme and $\mathcal{H}_{pr}$ is a classification head used to predict the category probability distribution of the pixel representations in $R$.
The matrix $C_{bi}$ with size of $Z \times \frac{H}{8} \times \frac{W}{8}$ denotes the dataset-level contextual information for each pixel.

To seamlessly integrate the proposed paradigm into the existing segmentation frameworks, we also define the self-existing context scheme within the individual images of the adopted framework as $\mathcal{A}_{wi}$. Then we have:
\begin{equation} \label{eq3}
   C_{wi} = \mathcal{A}_{wi}(R),
\end{equation}
where $C_{wi}$ is the contextual information within the input image.
Next, the original pixel representation $R$ is augmented with both the proposed dataset-level contextual information and the individual image contextual information as follows:
\begin{equation} \label{eq4}
   R_{aug} = \mathcal{F}_{T}(R, ~C_{bi}, ~C_{wi}),
\end{equation}
where $\mathcal{F}_{T}$ is a transform operator to fuse the original representations $R$, the contextual information beyond the input image $C_{bi}$ and the contextual information within the input image $C_{wi}$.
Note that $C_{wi}$ is optional according to the adopted segmentation framework. 
For example, if the adopted segmentation framework is FCN without any contextual modules, $R_{aug}$ will be calculated with $\mathcal{F}_{T}(R, ~C_{bi})$.
However, if the used segmentation framework is DeepLab, $C_{wi}$ is the output of ASPP.
Finally, the augmented pixel representation $R_{aug}$ is leveraged to generate the final semantic label of each pixel in the input image:
\begin{equation} \label{eq5}
   O = UP_{8\times}(\mathcal{H}_{cls} (R_{aug})),
\end{equation}
where $\mathcal{H}_{cls}$ is a classification head and $O$ is a matrix with size of $K \times H \times W$ and stores the semantic prediction of each pixel.
$K$ is the number of the predicted classes.

\subsection{Memory Module}
As illustrated in Fig.~\ref{framework}~(b), the memory module $\mathcal{M}$ with size of $K \times D$ is introduced to store the dataset-level distribution information of various classes, where $D$ denotes the number of hyperparameters used to represent the distribution.
In this work, we model the dataset-level category representations as a Gaussian distribution and set $D=2$ (\ie, mean and variance for each dataset-level category representation).
During training, we first randomly pick one pixel representation for each class from the train split and then compute its mean and standard deviation to initialize the memory module $\mathcal{M}$, where the representations are calculated by conducting $\mathcal{F}_{B}(I)$.
Then, the information in $\mathcal{M}$ are updated by conducting moving average after each training iteration:
\begin{equation} \label{eq6}
   \mathcal{M}_{t} = (1 - m) \cdot \mathcal{M}_{t-1} + m \cdot \mathcal{T}(R_{t-1}),
\end{equation}
where $t$ denotes for the current number of iterations and $m$ is the momentum. 
$\mathcal{T}(R_{t-1})$ is used to transform the feature representation $R_{t-1}$ to the distribution information of each category.
To obtain the distribution information from $R_{t-1}$, we first setup a matrix $\mathcal{M}'$ with the same size of $\mathcal{M}$, where the initial values in $\mathcal{M}'$ are copied from $\mathcal{M}$.
For convenience of the presentation, we use the subscript $[i, j]$ or $[i, *]$ to index the single element or elements in a certain position or row in the matrix.
$R$ is upsampled and permuted as size of $HW \times Z$, \ie, $R^{HW \times Z}$.
Then, for each category $c_k$ existing in the image $I$, we extract the feature representations of corresponding pixels:
\begin{equation} \label{eq7}
   R_{c_k} = \{R^{HW \times Z}_{[i, *]} ~|~  (GT_{[i]} = c_k) \land (1 \leq i \leq HW) \},
\end{equation}
where $GT$ represents the ground truth category labels.
After that, $R_{c_k} \in \mathbb{R}^{N_{c_k} \times C}$ stores the representations of category $c_k$, where $N_{c_k}$ is the number of pixels labeled as $c_k$ in the input image $I$.
Next, we calculate the average value of the pixel representations in $R_{c_k}$ as follows:
\begin{equation} \label{eq8}
   \hat{R}_{c_k} = \sum_{i=1}^{N_{c_k}} \frac{R_{c_k, [i, *]}}{N_{c_k}}.
\end{equation}
Finally, the distribution information of $c_k$ in $\mathcal{M}'$ is calculated as follows:
\begin{equation} \label{eq9}
\begin{aligned}
   \mathcal{M}'_{[c_k, 0]} &= \sum^{Z}_{i=1} \frac{\hat{R}_{c_k, [i]}}{Z},    \\
   \mathcal{M}'_{[c_k, 1]} &= \sqrt{\frac{1}{Z} \sum^{Z}_{i=1} (\hat{R}_{c_k, [i]} - \mathcal{M}'_{[c_k, 0]})^2}.
\end{aligned}
\end{equation}
$\mathcal{M}'$ will be updated with all the classes in the input image $I$ and is regarded as the output of $\mathcal{T}$ to update the memory module $\mathcal{M}$, following Eq.~\eqref{eq6}.

\subsection{Dataset-Level Context Aggregation} 
Here, we elaborate the dataset-level context aggregation scheme $\mathcal{A}_{bi}$ to adaptively aggregate the dataset-level category representations yielded by the dataset-level distribution information stored in $\mathcal{M}$.
As demonstrated in Fig.~\ref{framework}~(c), we first predict a weight matrix $W$ with a size of $K \times \frac{H}{8} \times \frac{W}{8}$ according to $R$:
\begin{equation} \label{eq10}
   W = \mathcal{H}_{pr}(R),
\end{equation}
where $W$ represents the category probability distribution of each pixel representation in $R$.
The classification head $\mathcal{H}_{pr}$ is implemented with two $1 \times 1$ convolutional layers following the Softmax function.
Meanwhile, we generate the dataset-level representations of various categories $C_{dl}$ according to the distribution information in $\mathcal{M}$:
\begin{equation} \label{eq11}
   C_{dl} = \mathcal{G}(\mathcal{M}),
\end{equation}
where $C_{dl}$ is with the size of $K \times Z$ and the generator $\mathcal{G}$ is a sampling operation from the Gaussian distribution determined by the parameters in $\mathcal{M}$.
The pseudo code of $\mathcal{G}$ is shown in Algorithm~\ref{alg:code}.

\begin{algorithm}
\caption{Pseudo code of generator $\mathcal{G}$.}
\label{alg:code}
\definecolor{codeblue}{rgb}{0.25,0.5,0.5}
\lstset{
  backgroundcolor=\color{white},
  basicstyle=\fontsize{7.2pt}{7.2pt}\ttfamily\selectfont,
  columns=fullflexible,
  breaklines=true,
  captionpos=b,
  commentstyle=\fontsize{7.2pt}{7.2pt}\color{codeblue},
  keywordstyle=\fontsize{7.2pt}{7.2pt},
  frame=none,
}
\begin{lstlisting}
# mean of each class
means = memory_module[:, 0]
# standard deviation of each class
stds = memory_module[:, 1]
# initialize dataset-level representations
dl_representations = []

# yield per-class dataset-level representation
for cls_idx in range(K):
    dl_representation[cls_idx] = torch.normal(
        mean=torch.full((1, Z), means[idx]), 
        std=torch.full((1, Z), stds[idx])
    )
\end{lstlisting}
\end{algorithm}

After obtaining the dataset-level category representations $C_{dl}$ and the weight matrix $W$, we calculate the coarse dataset-level representation for each pixel as follow:
\begin{equation} \label{eq12}
   \tilde{C}_{bi} = permute(W) \otimes C_{dl},
\end{equation}
where $\tilde{C}_{bi}$ is with the size of $\frac{HW}{64} \times Z$ and stores the aggregated dataset-level category representations for the pixel representations in $R$.
The $permute$ operation is used to arrange $W$ to the size of $\frac{HW}{64} \times K$ and $\otimes$ denotes the matrix multiplication.
Since $\mathcal{H}_{pr}$ only leverages the original pixel representation $R$ to predict $W$, while the pixel representations may be mis-classified.
Therefore, we further calculate the relation between $R$ and $\tilde{C}_{bi}$ to obtain a position confidence weight $P$ to further recalibrate $\tilde{C}_{bi}$ as follows:
\begin{equation} \label{eq13}
\begin{aligned}
   P &= Softmax(\frac{\mathcal{F}_{R_{1}}(permute(R)) \otimes \mathcal{F}_{R_{2}}(\tilde{C}_{bi})^T}{\sqrt{\frac{C}{2}}}),\\
   C_{bi} &= permute(\mathcal{F}_{O}(P \otimes \mathcal{F}_{C}(\tilde{C}_{bi}))),
\end{aligned}
\end{equation}
where the $permute(R)$ operation is adopted to arrange $R$ to the size of $\frac{HW}{64} \times Z$.
$\mathcal{F}_{R_{1}}$, $\mathcal{F}_{R_{2}}$, $\mathcal{F}_{O}$ and $\mathcal{F}_{C}$ are introduced to adjust the dimension of each pixel representation and they are implemented by a $1 \times 1$ convolutional layer.
The last $permute$ operation is used to resize the output to have the size of $Z \times \frac{H}{8} \times \frac{W}{8}$.

\begin{figure}
\centering
\includegraphics[width=0.48\textwidth]{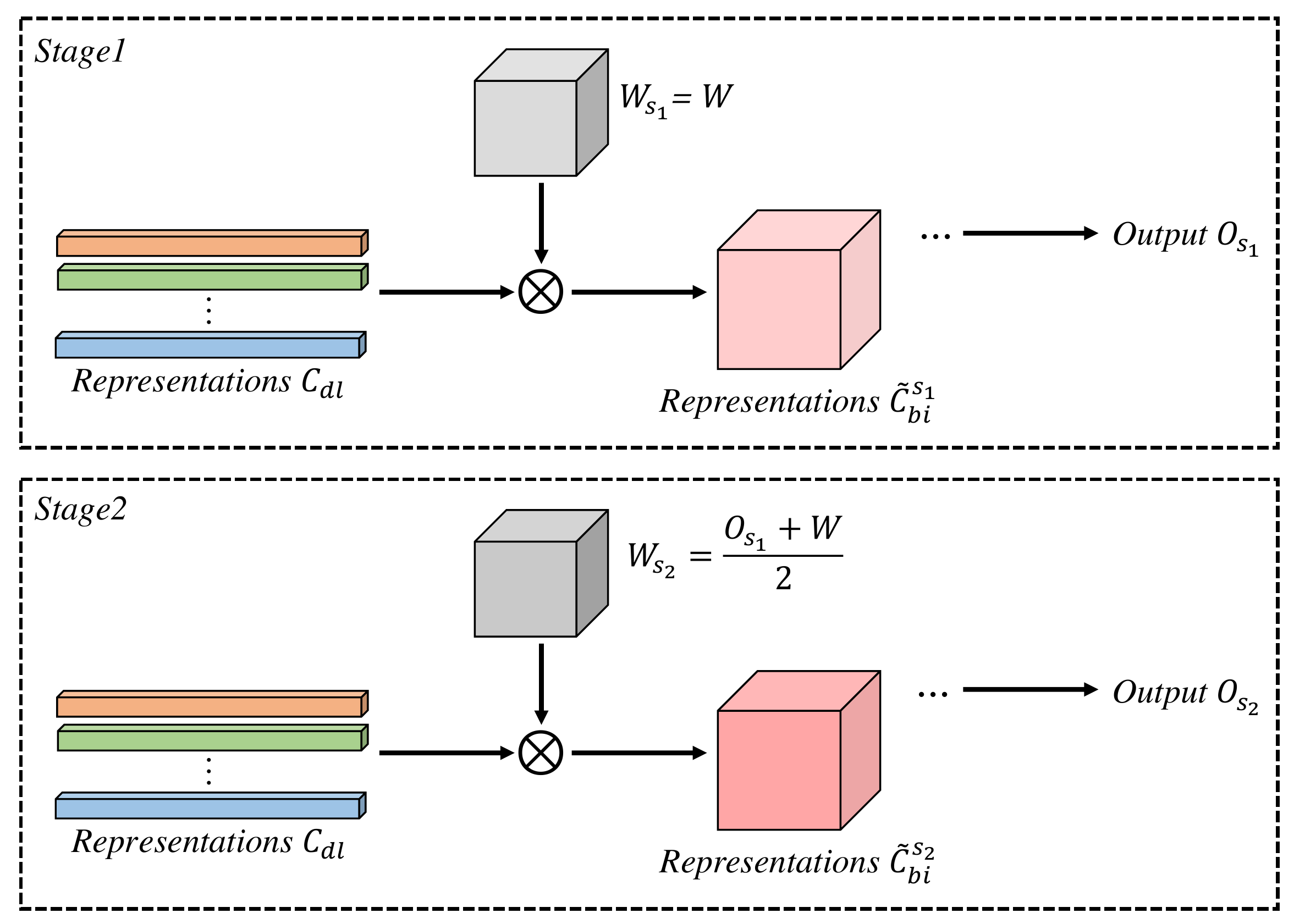}
\caption{
   Illustration of the proposed coarse-to-fine iterative inference strategy.
   $\otimes$ denotes the matrix multiplication.
   Dataset-level representations $C_{dl}$ are shared in \emph{Stage1} and \emph{Stage2}.
}\label{figTestingStrategy}
\vspace{-0.40cm}
\end{figure}

\subsection{Objective Functions} 
A joint loss function of the weight matrix $W$ and the final prediction $O$ is used to optimize the model parameters.
Specifically, the objective of $W$ is defined as:
\begin{equation} \label{eq14}
   \mathcal{L}_{W} = \frac{1}{H \times W} \sum_{i, j} CE (W'_{[*, i,j]}, ~GT_{[ij]}),
\end{equation}
where $W'$ is the upsampled prediction $UP_{8 \times}(W)$.
$CE$ denotes for the cross entropy loss and $\sum_{i, j}$ denotes that the summation is calculated over all locations on the input image $I$.
Moreover, to make the final segmentation prediction $O$ contain the accurate category probability distribution of each pixel, we define the objective of $O$ as:
\begin{equation} \label{eq15}
   \mathcal{L}_{O} = \frac{1}{H \times W} \sum_{i, j} CE (O_{[*, i, j]}, ~GT_{[ij]}).
\end{equation}
Finally, we formulate the joint loss function $\mathcal{L}$ to train the whole segmentation framework as follows:
\begin{equation} \label{eq16}
   \mathcal{L} = \alpha \mathcal{L}_{W} + \mathcal{L}_{O},
\end{equation}
where $\alpha$ is the hyper-parameter to balance the two losses. 
Following \cite{jin2021mining}, we empirically set $\alpha=0.4$ by default.
With this joint loss function, the model parameters are updated jointly through back propagation.
Note that the values in $\mathcal{M}$ are not learnable through the back propagation.

\subsection{Coarse-to-fine Iterative Inference Strategy} 
In the inference phase, we develop a coarse-to-fine iterative inference strategy to progressively incorporate accurate dataset-level representation for each pixel.
As indicated in Fig.~\ref{figTestingStrategy}, in \emph{Stage1}, we first conduct Eq.~(\ref{eq10})-(\ref{eq12}) to gather the first-stage dataset-level representations $\tilde{C}^{s_{1}}_{bi}$ for the pixels.
We employ $\tilde{C}^{s_{1}}_{bi}$ to augment the original pixel representations $R$ and obtain the first-stage category probability distribution of each pixel, \ie, $O_{s_{1}}$.
During \emph{Stage2}, we employ the first-stage prediction $O_{s_{1}}$ to update the weights for dataset-level context aggregation. 
Specifically, we calculate the weights in \emph{Stage2} as:
\begin{equation} \label{eq17}
   W_{s_{2}} = \frac{O_{s_{1}} + W}{2}.
\end{equation}
After that, we replace $W_{s_{1}}$ (\ie, $W$ predicted by Eq.(\ref{eq10})) with $W_{s_{2}}$ and re-conduct the dataset-level context aggregation as Eq.(\ref{eq10})-(\ref{eq12}) to obtain $\tilde{C}^{s_{2}}_{bi}$.
Finally, we utilize the pixel representations augmented by $\tilde{C}^{s_{2}}_{bi}$ to predict the final class probability distribution of each pixel, \ie, $O_{s_{2}}$.
As the predicted category probability distribution in $O_{s_1}$ is more accurate than $W$, $W_{s_2}$ can help to improve the accuracy of the aggregated dataset-level category representations for the pixels and thereby achieve better segmentation results.

Following the steps above, we can also obtain $W_{s_{i}},~i \geq 2$ to further boost the segmentation performance. 
In particular, we set $i=2$ by default for best trade-off.

\begin{figure*}[t]
\centering
\includegraphics[width=1.0\textwidth]{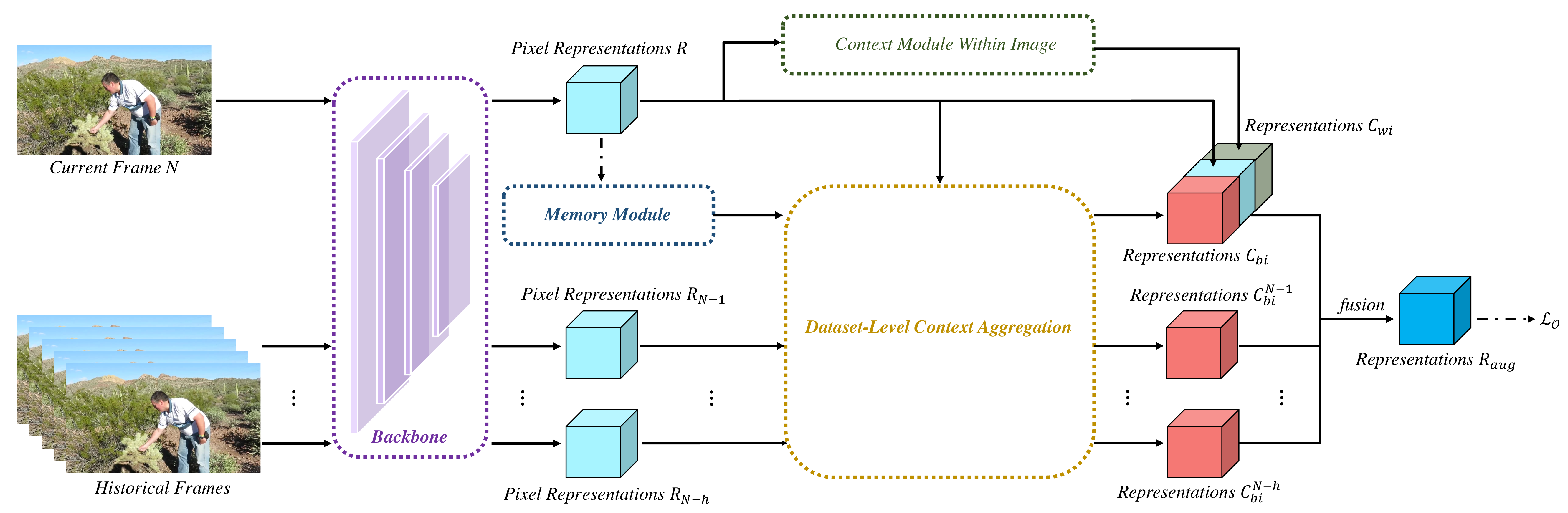}
\vspace{-0.40cm}
\caption{
   Extending soft mining contextual information beyond the input image (MCIBI++) into the video semantic segmentation task.
   The historical frames are inputted to the proposed framework to obtain the temporal information for the segmentor.
}\label{frameworkvideo}
\vspace{-0.40cm}
\end{figure*}

\subsection{Extension to Video Semantic Segmentation} \label{method:VSS}
In this subsection, we extend MCIBI++ to the video semantic segmentation task.
The key idea of the extension is to introduce the extra temporal information by leveraging the memory module.
Accordingly, the image semantic segmentation framework equipped with MCIBI++ can be effectively and conveniently extent to the temporal dimension.

\para{Baseline Framework.}
Although video semantic segmentation is different from image semantic segmentation, a stronger image semantic segmentor can usually achieve better performance in the video semantic segmentation.
Therefore, we adopt the image semantic segmentation framework equipped with MCIBI++ as our baseline framework.
Specifically, we first perform Eq.~(\ref{eq1})-(\ref{eq3}) to obtain $R$, $C_{bi}$ and $C_{wi}$.
For our implementation, the adopted backbone networks are ResNet-101 and Swin-Large \cite{liu2021swin} and the employed context module within image is PPM plus FPN proposed in UperNet~\cite{xiao2018unified}.

\para{Incorporating Temporal Information.}
As demonstrated in Fig.~\ref{frameworkvideo}, we also feed the historical frames into the network to introduce the temporal information for our video semantic segmentation framework.
In details, given the historical frame $I_{N-i},1 \leq i \leq h$, we first adopt the backbone network to extract the historical pixel representations $R_{N-i}$.
Then, we calculate the aggregation weights for these historical pixel representations as follow:
\begin{equation} \label{eq18}
   W_{N-i} = \mathcal{H}_{pr}(R_{N-i}),~1 \leq i \leq h,
\end{equation}
where $h$ is the number of historical frames. 
In our implementation, we set $h=2$.
Then, we leverage the calculated aggregation weights $W_{N-i}$ to gather the dataset-level representations for each historical frame by performing Eq.~(\ref{eq11})-(\ref{eq13}) and thereby we can obtain $C_{bi}^{N-i},1 \leq i \leq h$.
Finally, we combine them together to obtain the augmented pixel representation:
\begin{equation} \label{eq19}
   R_{aug} = \mathcal{F}_{T}(R, C_{wi}, C_{bi}, C_{bi}^{N-1}, ..., C_{bi}^{N-h}),
\end{equation}
where $R_{aug}$ is as the input of Eq.~(\ref{eq5}) to predict the category probability distribution of each pixel in $I_{N}$.

\para{Discussion.}
In fact, there are other strategies to exploit the temporal information from the historical frames.
For example, $C_{bi}^{N-i}$ in Eq.~(\ref{eq19}) can also be replaced with $R_{N-i}$, which seems more convenient to be implemented.
However, according to our experience, introducing $C_{wi}$ and $C_{bi}$ to $R_{aug}$ can bring more performance gains for video semantic segmentation than the direct temporal information $R_{N-i}$.
If we adopt $R_{N-i}$ in Eq.~(\ref{eq19}), the proportion of the contextual information will decease with the increase of the utilized historical frames, which may suppress the segmentation performance.
In addition, the appearance of the same object may vary a lot in different frames and thus the pixel representations in $R_{N-i}$ for the same object may also vary a lot during training, 
which is not friendly for the network to capture temporal relations of the objects between the frames.
This problem can be alleviated when adopting $C_{bi}^{N-i}$, as the dataset-level representation for each object keeps consistent during different frames.
\section{Experiments on Image Semantic Segmentation}
We first present an extensive evaluation of our framework on the image semantic segmentation task.

\subsection{Experimental Setup}
\para{Datasets.}
We conduct experiments on four widely-used semantic segmentation benchmarks:
\begin{itemize}
   \item \textbf{ADE20K}~\cite{zhou2017scene} is a challenging scene parsing dataset that contains 150 classes and diverse scenes with 1,038 image-level labels.
   The training, validation, and test sets consist of 20K, 2K, 3K images, respectively.
   \item \textbf{LIP}~\cite{gong2017look} is tasked for single human parsing. There are 19 semantic classes and one background class.
   The dataset is divided into 30K/10K/10K images for training, validation and testing, respectively.
   \item \textbf{Cityscapes}~\cite{cordts2016cityscapes} densely annotates 19 object categories in urban scene understanding images.
   This dataset consists of 5,000 finely annotated images, where 2,975 are available for training, 500 for validation and  1,524 for testing.
   \item \textbf{COCO-Stuff} \cite{caesar2018coco} is a challenging scene parsing dataset that provides rich annotations for 91 thing classes and 91 stuff classes. 
   This work adopts the old version which includes 10K images (\ie, 9K for training and 1K for testing) from MS COCO \cite{lin2014microsoft}.
\end{itemize}

\para{Training Settings.} 
Our backbone network was initialized by the weights pre-trained on ImageNet \cite{deng2009imagenet}, while the remaining layers were randomly initialized. 
Following the previous work~\cite{chen2017deeplab}, the learning rate was decayed according to the ``poly'' learning rate policy with factor $(1 - \frac{iter}{total\_iter})^{0.9}$.
Synchronized batch normalization implemented by pytorch was enabled during training.
For data augmentation, we adopted random scaling, horizontal flipping and color jitter.
More specifically, the training settings for different datasets are listed as follows:
\begin{itemize}
   \item \textbf{ADE20K:} 
   We set the initial learning rate as $0.01$,
   weight decay as $0.0005$, 
   crop size as $512 \times 512$, 
   batch size as $16$ and 
   training epochs as $130$ by default.
   \item \textbf{LIP:} 
   The initial learning rate was set as $0.01$ and the weight decay was $0.0005$. 
   We set the crop size of the input image as $473 \times 473$ and batch size as $32$ by default.
   Besides, if not specified, the networks were fine-tuned for $150$ epochs on the train set.
   \item \textbf{Cityscapes:}    
   The initial learning rate was set as $0.01$ and the weight decay was $0.0005$.
   We randomly cropped out high-resolution patches $512 \times 1024$ from the original images as the inputs for training.
   The batch size and training epochs were set as $8$ and $220$, respectively.
   \item \textbf{COCO-Stuff:} 
   We set the initial learning rate as $0.001$,
   weight decay as $0.0001$, 
   crop size as $512 \times 512$, 
   batch size as $16$ and 
   training epochs as $110$ by default.
\end{itemize}

\noindent \textbf{Inference Settings.} 
For ADE20K, COCO-Stuff and LIP, the size of the input image during testing was the same as the size of the input image during training.
And for Cityscapes, the input image was resized to have its shorter side being $1024$ pixels. 
By default, no tricks (\eg, multi-scale with flipping testing) were used during testing.

\noindent \textbf{Evaluation Metrics.} 
Following the standard setting, mean intersection-over-union (mIoU) was adopted for evaluation.

\noindent \textbf{Reproducibility.} 
Our method was implemented with PyTorch ($version \geq 1.3$) \cite{paszke2019pytorch}
and was trained on eight NVIDIA Tesla V100 GPUs with a 32 GB memory per-card.
All the testing procedures were performed on a single NVIDIA Tesla V100 GPU.
To provide full details of our framework, our code will be made publicly available in \cite{ssseg2020}.

\subsection{Ablation Analysis} \label{ImageSegmentorAblationAnalysis}

\para{Transform Function $\mathcal{F}_{T}$.}
Table~\ref{fusionfunction} shows the performance of three fusion strategies to fuse $R$ and $C_{bi}$ on the validation set of ADE20K.
Specifically, \emph{add} denotes for element-wise adding, which brings $4.19\%$ mIoU improvements compared with the baseline model FCN.
\emph{weighted add} indicates weighted summation of $R$ and $C_{bi}$, where the weights are predicted by a $1 \times 1$ convolutional layer.
It achieves $5.10\%$ mIoU improvements over the baseline model and outperforms the naive summation. 
\emph{concatenation} stands for concatenating $R$ and $C_{bi}$ and achieves the optimal performance among three strategies. 
All the three strategies achieve the consistent improvements over the baseline model, indicating that the proposed MCIBI++ paradigm can effectively improve the pixel representations so that the networks can classify the pixels more accurately.

\para{Memory Module.} 
Table~\ref{ablation} shows the ablation study of the memory module $\mathcal{M}$ on the validation set of ADE20K.
Comparing the 1st and 4th rows, we can observe that leveraging $\mathcal{M}$ to introduce the contextual information beyond the input image can boost the performance of baseline segmentation framework (from $36.96\%$ to $43.39\%$ mIoU).
This result indicates that the yielded dataset-level representations can effectively improve the original pixel representations.
In addition, it is also observed that MCIBI++ (4th row) outperforms our previous MCIBI~\cite{jin2021mining} (3rd row) by $1.21\%$, 
showing that storing the dataset-level distribution information of each category instead of directly storing the feature representations can obtain more discriminative dataset-level cateogory representations.
Furthermore, the segmentation framework with non-updated $\mathcal{M}$ achieves poor performance (2nd row, only hitting $21.78\%$ mIoU), 
indicating that one key success factor of $\mathcal{M}$ is the dynamic update mechanism to maintain the suitable dataset-level distribution information of various classes in the memory module. 

We further conduct the ablation studies in Table~\ref{momentum} to find the most effective value of momentum $m$.
It is observed that setting $m=0.1$ achieve the optimal performance. 
Moreover, although adopting the polynomial annealing policy to schedule $m$ seems more flexible and has been adopted in \cite{jin2021mining}, 
we observe that using a fixed momentum value brings the better segmentation results after replacing the dataset-level category representations with the dataset-level distribution information in $\mathcal{M}$.

\begin{table}[t]
\centering
\caption{
   Ablation study of $\mathcal{F}_{T}$ on ADE20K val split.
}\label{fusionfunction}
\vspace{-0.3cm}
\resizebox{.48\textwidth}{!}{
\begin{tabular}{c|c|c|c}
   \hline
   \hline
   Method                                &Backbone      &$\mathcal{F}_{T}$       &mIoU ($\%$)      \\
   \hline
   FCN                                   &ResNet-50     &-                       &36.96            \\
   \hline
   FCN+MCIBI++ (\emph{ours})             &ResNet-50     &\emph{add}              &41.15            \\
   FCN+MCIBI++ (\emph{ours})             &ResNet-50     &\emph{weighted add}     &42.06            \\
   FCN+MCIBI++ (\emph{ours})             &ResNet-50     &\emph{concatenation}    &\textbf{43.39}   \\
   \hline
   \hline
\end{tabular}}
\end{table}

\begin{table}[t]
\centering
\caption{
   Ablation study of MCIBI++ on ADE20K val split.
   Memory-R means directly storing the dataset-level representations of categories in $\mathcal{M}$, 
   while Memory-D denotes maintaining the distribution information of classes in $\mathcal{M}$.
   Update indicates whether to update the memory module during training and IIS refers to the proposed iterative inference strategy.
}\label{ablation}
\vspace{-0.3cm}
\resizebox{.48\textwidth}{!}{
\begin{tabular}{ccccc|c}
   \hline
   \hline
   FCN              &Memory-R          &Memory-D       &Update           &IIS               &mIoU ($\%$)      \\
   \hline
   \checkmark       &                  &               &                 &                  &36.96            \\
   \checkmark       &\checkmark        &               &                 &                  &21.78            \\
   \checkmark       &\checkmark        &               &\checkmark       &                  &42.18            \\
   \checkmark       &                  &\checkmark     &\checkmark       &                  &43.39            \\
   \checkmark       &                  &\checkmark     &\checkmark       &\checkmark        &\textbf{43.74}   \\
   \hline
   \hline
\end{tabular}}
\vspace{-0.3cm}
\end{table}

\begin{table}[t]
\centering
\caption{
   Ablation study of momentum $m$ of $\mathcal{M}$ on ADE20K val set.
   $0.9$-Poly means adopting the polynomial annealing policy to schedule $m$~\cite{jin2021mining}.
}\label{momentum}
\vspace{-0.3cm}
\resizebox{.48\textwidth}{!}{
\begin{tabular}{c|c|c|c}
   \hline
   \hline
   Momentum $m$                   &Backbone      &Stride        &mIoU ($\%$)       \\
   \hline        
   $0.01$                         &ResNet-50     &$8\times$     &42.06             \\
   $0.05$                         &ResNet-50     &$8\times$     &42.89             \\
   $0.1$                          &ResNet-50     &$8\times$     &\textbf{43.39}    \\
   $0.3$                          &ResNet-50     &$8\times$     &42.71             \\
   $0.5$                          &ResNet-50     &$8\times$     &42.72             \\
   $0.7$                          &ResNet-50     &$8\times$     &43.23             \\
   $0.9$                          &ResNet-50     &$8\times$     &42.95             \\
   $0.9$-Poly                     &ResNet-50     &$8\times$     &42.45             \\
   \hline
   \hline
\end{tabular}}
\end{table}

\begin{table}[t]
\centering
\caption{
   Ablation study of loss weight $\alpha$ on ADE20K val set.
}\label{lossweight}
\vspace{-0.3cm}
\resizebox{.48\textwidth}{!}{
\begin{tabular}{c|c|c|c}
   \hline
   \hline
   Loss Weight $\alpha$           &Backbone      &Stride        &mIoU ($\%$)       \\
   \hline        
   $0.2$                          &ResNet-50     &$8\times$     &43.35             \\
   $0.4$                          &ResNet-50     &$8\times$     &\textbf{43.39}    \\
   $0.6$                          &ResNet-50     &$8\times$     &43.25             \\
   $0.8$                          &ResNet-50     &$8\times$     &42.83             \\
   $1.0$                          &ResNet-50     &$8\times$     &42.22             \\
   \hline
   \hline
\end{tabular}}
\vspace{-0.30cm}
\end{table}

\para{Loss Weight.} 
Table~\ref{lossweight} shows the ablation study of the loss weight $\alpha$ on the validation set of ADE20K.
We can see that setting $\alpha=0.4$ achieve the best segmentation performance.

\begin{table}[t]
\centering
\caption{
   Ablation study of the number of the stages of IIS on ADE20K val set.
   Trade-off is obtained by dividing the improved mIoU by the increased Time.
   All the numbers are obtained on a single NVIDIA A100 Tensor Core GPU.
}\label{iistimes}
\vspace{-0.3cm}
\resizebox{.48\textwidth}{!}{
\begin{tabular}{c|c|c|c|c|c}
   \hline
   \hline
   Stage $i$       &Backbone      &Stride        &mIoU ($\%$)       &Time (ms)  &Trade-off            \\
   \hline        
   1               &ResNet-50     &$8\times$     &43.39             &34.91      &-                    \\
   2               &ResNet-50     &$8\times$     &43.74             &49.33      &\textbf{0.024}       \\
   3               &ResNet-50     &$8\times$     &43.85             &58.19      &0.012                \\
   4               &ResNet-50     &$8\times$     &43.91             &74.72      &0.004                \\
   \hline
   \hline
\end{tabular}}
\end{table}

\begin{table}[t]
\centering
\caption{
   Complexity comparison. 
   The input feature maps of the context modules are of size $[1 \times 2048 \times 128 \times 128]$ to evaluate their complexity during inference.
   All the numbers are obtained on a single NVIDIA Tesla V100 GPU. 
   Except for mIoU, the smaller values indicate the better performance.
}\label{complexity}
\vspace{-0.30cm}
\resizebox{.48\textwidth}{!}{
\begin{tabular}{c|c|c|c|c|c}
   \hline
   \hline
   Method                                              &Params          &FLOPs            &Time                &Memory          &mIoU ($\%$)    \\
   \hline
   ASPP (\emph{our impl.})                             &42.21M          &674.47G          &101.44ms            &1.68G           &43.19          \\
   PPM (\emph{our impl.})                              &23.07M          &309.45G          &\textbf{29.57ms}    &\textbf{0.97G}  &42.64          \\
   CCNet (\emph{our impl.})                            &23.92M          &397.38G          &56.90ms             &1.27G           &42.47          \\
   DANet (\emph{our impl.})                            &23.92M          &392.02G          &62.64ms             &5.16G           &42.90          \\
   ANN (\emph{our impl.})                              &20.32M          &335.24G          &49.66ms             &2.28G           &41.75          \\
   DNL (\emph{our impl.})                              &24.12M          &395.25G          &68.62ms             &5.02G           &43.50          \\
   APCNet (\emph{our impl.})                           &30.46M          &413.12G          &54.20ms             &2.03G           &43.47          \\
   \hline
   \textbf{\emph{MCIBI}}&&&&& \\
   DCA (\emph{ours})                                   &\textbf{14.82M} &\textbf{242.80G} &47.64ms             &3.65G           &42.18           \\
   ASPP+DCA (\emph{ours})                              &45.63M          &730.56G          &125.03ms            &5.28G           &44.34           \\
   \hline
   \textbf{\emph{MCIBI++}}&&&&& \\
   DCA (\emph{ours})                                   &\textbf{14.82M} &\textbf{242.80G} &44.45ms             &2.96G           &43.39           \\
   ASPP+DCA (\emph{ours})                              &45.63M          &730.56G          &112.64ms            &4.59G           &\textbf{44.85}  \\
   \hline
   \hline
\end{tabular}}
\vspace{-0.30cm}
\end{table}

\begin{table}[t]
\centering
\caption{
   The generalization ability of MCIBI++ on other test set.
   LIP and CIHP contain exactly the same semantic labels.
}\label{generalization}
\vspace{-0.30cm}
\resizebox{.48\textwidth}{!}{
\begin{tabular}{c|c|c|c}
   \hline
   \hline
   Method                                &Train Set           &Test Set         &mIoU ($\%$) \\
   \hline
   FCN (\emph{our impl.})                &LIP train set       &LIP val set      &48.63       \\
   UperNet (\emph{our impl.})            &LIP train set       &LIP val set      &52.95       \\
   FCN (\emph{our impl.})                &LIP train set       &CIHP val set     &27.20       \\
   UperNet (\emph{our impl.})            &LIP train set       &CIHP val set     &26.84       \\
   \hline
   FCN+MCIBI++ (\emph{ours})             &LIP train set       &LIP val set      &51.13 (\textbf{{\color{teal}+2.50}}) \\
   UperNet+MCIBI++ (\emph{ours})         &LIP train set       &LIP val set      &53.92 (\textbf{{\color{teal}+0.97}}) \\
   FCN+MCIBI++ (\emph{ours})             &LIP train set       &CIHP val set     &27.57 (\textbf{{\color{teal}+0.37}}) \\
   UperNet+MCIBI++ (\emph{ours})         &LIP train set       &CIHP val set     &27.74 (\textbf{{\color{teal}+0.90}}) \\
   \hline
   \hline
\end{tabular}}
\vspace{-0.30cm}
\end{table}

\para{Coarse-to-fine Iterative Inference Strategy.}
In our preliminary experiments as shown in Table~\ref{upperbound}, we find that improving the accuracy of aggregation weights $W$ can boost the segmentation performance.
Therefore, we propose a coarse-to-fine iterative inference strategy to improve the segmentation results in the inference phase.
From Table~\ref{ablation}, we can observe that the designed iterative inference strategy brings $0.35\%$ improvements on mIoU (comparing 4th and 5th rows), demonstrating the effectiveness of our strategy.
To further validate this strategy, we also calculate the mIoU and accuracy of the aggregation weights $W$ at different stages.
Particularly, $W_{s_2}$ obtains $40.89\%$ mIoU and $79.40\%$ pixel accuracy while $W_{s_1}$ only achieves $36.44\%$ mIoU and $77.03\%$ pixel accuracy, which is along the lines of our motivation.

We further conduct ablation experiments in Table \ref{iistimes} to investigate the most effective number of the stages in the coarse-to-fine iterative inference strategy.
From this table, it is observed that setting stage $i=2$ can obtain the better trade-off between accuracy improvement and computation cost.
(Note that stage $i = 1$ denotes for the baseline method without IIS.)

\begin{table*}[t]
\centering
\caption{
   The segmentation performance improvements on various benchmarks when integrating MCIBI++ into four existing segmentation frameworks. 
}\label{improvements}
\vspace{-0.3cm}
\resizebox{.95\textwidth}{!}{
\begin{tabular}{c|c|c|c|c|c|c}
   \hline
   \hline
   Method                                                   &Backbone      &Stride     &ADE20K (\emph{train}/\emph{val})       &Cityscapes (\emph{train}/\emph{val})      &LIP (\emph{train}/\emph{val})         &COCO-Stuff (\emph{train}/\emph{test})          \\
   \hline
   FCN (\emph{our impl.})                                   &ResNet-50     &$8\times$  &36.96                                  &75.16                                     &48.63                                 &31.76                                          \\
   FCN+MCIBI++ (\emph{ours})                                &ResNet-50     &$8\times$  &43.39 (\textbf{{\color{teal}+6.43}})   &78.77 (\textbf{{\color{teal}+3.61}})      &51.13 (\textbf{{\color{teal}+2.50}})  &37.38 (\textbf{{\color{teal}+5.62}})           \\
   \hline
   PSPNet (\emph{our impl.})                                &ResNet-50     &$8\times$  &42.64                                  &79.05                                     &51.94                                 &37.40                                          \\
   PSPNet+MCIBI++ (\emph{ours})                             &ResNet-50     &$8\times$  &43.89 (\textbf{{\color{teal}+1.25}})   &79.91 (\textbf{{\color{teal}+0.86}})      &52.93 (\textbf{{\color{teal}+0.99}})  &38.47 (\textbf{{\color{teal}+1.07}})           \\
   \hline
   UperNet (\emph{our impl.})                               &ResNet-50     &$8\times$  &43.02                                  &79.08                                     &52.95                                 &37.65                                          \\
   UperNet+MCIBI++ (\emph{ours})                            &ResNet-50     &$8\times$  &44.30 (\textbf{{\color{teal}+1.28}})   &80.05 (\textbf{{\color{teal}+0.97}})      &53.92 (\textbf{{\color{teal}+0.97}})  &39.20 (\textbf{{\color{teal}+1.55}})           \\
   \hline
   DeepLabV3 (\emph{our impl.})                             &ResNet-50     &$8\times$  &43.19                                  &79.62                                     &52.35                                 &37.63                                          \\
   DeepLabV3+MCIBI++ (\emph{ours})                          &ResNet-50     &$8\times$  &44.85 (\textbf{{\color{teal}+1.66}})   &80.72 (\textbf{{\color{teal}+1.10}})      &53.59 (\textbf{{\color{teal}+1.24}})  &38.94 (\textbf{{\color{teal}+1.31}})           \\
   \hline
   ISNet (\emph{our impl.})                                 &ResNet-50     &$8\times$  &44.22                                  &79.32                                     &53.14                                 &38.06                                          \\
   ISNet+MCIBI++ (\emph{ours})                              &ResNet-50     &$8\times$  &45.33 (\textbf{{\color{teal}+1.11}})   &80.49 (\textbf{{\color{teal}+1.17}})      &54.06 (\textbf{{\color{teal}+0.92}})  &39.33 (\textbf{{\color{teal}+1.27}})           \\
   \hline
   \hline
\end{tabular}}
\vspace{-0.3cm}
\end{table*}

\begin{table}[t]
\centering
\caption{
   The comparison with the state-of-the-art methods on ADE20K (val).
   Multi-scale and flipping testing are adopted for fair comparison.
   \textbf{Bold} denotes the best score.
}\label{sotaade20k}
\vspace{-0.3cm}
\resizebox{.48\textwidth}{!}{
\begin{tabular}{c|c|c|c}
   \hline
   \hline
   Method                                           &Backbone      &Stride                  &mIoU ($\%$)           \\
   \hline
   CCNet \cite{huang2019ccnet}                      &ResNet-101    &$8\times$               &45.76                 \\
   OCNet \cite{yuan2018ocnet}                       &ResNet-101    &$8\times$               &45.45                 \\
   ACNet \cite{fu2019adaptive}                      &ResNet-101    &$8\times$               &45.90                 \\
   DMNet \cite{he2019dynamic}                       &ResNet-101    &$8\times$               &45.50                 \\
   EncNet \cite{zhang2018context}                   &ResNet-101    &$8\times$               &44.65                 \\
   PSPNet \cite{zhao2017pyramid}                    &ResNet-101    &$8\times$               &43.29                 \\
   PSANet \cite{zhao2018psanet}                     &ResNet-101    &$8\times$               &43.77                 \\
   APCNet \cite{he2019adaptive}                     &ResNet-101    &$8\times$               &45.38                 \\
   OCRNet \cite{yuan2019object}                     &ResNet-101    &$8\times$               &45.28                 \\
   UperNet \cite{xiao2018unified}                   &ResNet-101    &$8\times$               &44.85                 \\
   CPNet \cite{yu2020context}                       &ResNet-101    &$8\times$               &46.27                 \\
   ISNet \cite{jin2021isnet}                        &ResNet-101    &$8\times$               &47.31                 \\
   MaskFormer \cite{cheng2021per}                   &ResNet-101    &$8\times$               &47.20                 \\
   PSPNet \cite{zhao2017pyramid}                    &ResNet-269    &$8\times$               &44.94                 \\
   OCRNet \cite{yuan2019object}                     &HRNetV2-W48   &$4\times$               &45.66                 \\
   ISNet \cite{jin2021isnet}                        &ResNeSt-101   &$8\times$               &47.55                 \\
   DeepLabV3 \cite{chen2017rethinking}              &ResNeSt-101   &$8\times$               &46.91                 \\
   DeepLabV3 \cite{chen2017rethinking}              &ResNeSt-200   &$8\times$               &48.36                 \\
   SETR-MLA \cite{zheng2021rethinking}              &ViT-Large     &$16\times$              &50.28                 \\
   UperNet \cite{liu2021swin}                       &Swin-Large    &$32\times$              &53.50                 \\
   \hline
   DeepLabV3+MCIBI \cite{jin2021mining}             &ResNet-101    &$8\times$               &47.22                 \\
   DeepLabV3+MCIBI \cite{jin2021mining}             &ResNeSt-101   &$8\times$               &47.36                 \\
   DeepLabV3+MCIBI \cite{jin2021mining}             &ViT-Large     &$16\times$              &50.80                 \\
   \hline
   UperNet+MCIBI++ (\emph{ours})                    &ResNet-101    &$8\times$               &47.93                 \\
   UperNet+MCIBI++ (\emph{ours})                    &ResNeSt-101   &$8\times$               &48.56                 \\
   UperNet+MCIBI++ (\emph{ours})                    &Swin-Large    &$32\times$              &\textbf{54.52}        \\
   \hline
   \hline
\end{tabular}}
\vspace{-0.3cm}
\end{table}

\para{The Initialization of $\mathcal{M}$.}
By default, each item in $\mathcal{M}$ is initialized by the $mean$ and $std$ of a randomly selecting pixel representation of the corresponding category.
To test the influence of the initialization method on the segmentation performance, we also initialize $\mathcal{M}$ with the $mean$ and $std$ of the composite vector of each category,
where the composite vectors are obtained by cosine similarity clustering.
The performance of this strategy is comparable with the random initialization ($43.43\%$ \emph{v.s.} $43.39\%$), which is consistent with the observation in MCIBI~\cite{jin2021mining}.
This comparison indicates that the dynamic updating mechanism can make $\mathcal{M}$ converge and without relying on a strict initialization, which shows the robustness of our method.

\para{Complexity Comparison.} 
In Table \ref{complexity}, we show the efficiency of the proposed dataset-level context aggregation (DCA) scheme.
Compared with the existing context schemes, our DCA requires the least computation complexity ($242.80$ GFLOPs) and the least parameters ($14.82$ M), showing that the proposed context module is light.
Since DCA aims to aggregate the contextual information beyond the input image while the existing context schemes focus on gathering contextual information within the image, DCA is complementary to the existing context schemes.
The lightweight attribute of DCA makes the whole model complexity still tolerable after integrating the proposed module with other existing within image context schemes to further boost performance.
For example, after integrating DCA with ASPP, the parameters, FLOPs and inference time just increase by $3.42$M, $56.09$G and $11.20$ms, respectively (comparing the first row and the last row in Table \ref{complexity}), 
while the segmentation performance boosts from $43.19\%$ to $44.85\%$.
This comparison well validate our motivation, \ie, the contexts beyond image $C_{bi}$ and the contexts within image $C_{wi}$ are complementary.
It is worth noting that only incorporating the MCIBI++ paradigm can also boost the baseline FCN framework to reach a $43.39\%$ mIoU, which is higher than most of the existing context schemes.

\para{Training Memory.}
Table~\ref{complexity} also compares the maximum training memory required for MCIBI and MCIBI++.
It is observed that the training memory decreases from $3.65$G to $2.96$G after replacing representations with distribution information in the memory module.
The drop is mainly due to the reduction of the memory required to update the memory module $\mathcal{M}$.
Specifically, during training, MCIBI leverages the feature representations stored in GPU to update $\mathcal{M}$ while MCIBI++ only utilizes the distribution information to update $\mathcal{M}$.
Obviously, the dimension of the tensor storing distribution information is much smaller than that storing feature representations.
Therefore, the proposed MCIBI++ can benefit in the saving of GPU memory during network training compared with our previous MCIBI.
Note that, in the inference phase, the GPU memory requirements for MCIBI and MCIBI++ will decrease a lot since the memory module on GPU is without updating.

\para{Overfitting.} 
As the memory module first stores the dataset-level distribution information of different categories and then generate the dataset-level contextual information to augment the original pixel representations, 
this memory module may lead the segmentation framework to overfit on the current dataset.
To investigate this point, we train the baseline segmentors (\ie, FCN and UperNet) and our proposed framework on the train set of LIP, and then evaluate them on the validation set of both LIP and CIHP \cite{gong2018instance}, where CIHP is a much more challenging multi-human parsing dataset with the same categories as LIP and can be used to evaluate the generalization ability of the networks.
As demonstrated in Table~\ref{generalization}, our framework consistently boosts the segmentation performance of two baseline segmentors on both LIP val set ($51.13\%$ \emph{v.s.} $48.63\%$ and $53.92\%$ \emph{v.s.} $52.95\%$ for FCN and UperNet, respectively) and CIHP val set ($27.57\%$ \emph{v.s.} $27.20\%$ and $27.74\%$ \emph{v.s.} $26.84\%$).
These comparisons indicate that the distribution information of LIP also benefits the segmentation on CIHP dataset and introducing the memory module can help the segmentors improve the generalization ability to some extent.

\subsection{Integrated into Various Segmentation Frameworks} 
Our MCIBI++ is orthogonal to the segmentation network design and it can be seamlessly incorporated into existing segmentation frameworks.
Therefore, we integrate the proposed MCIBI++ into five existing segmentation frameworks, \ie, FCN, PSPNet, UperNet, DeepLabV3 and ISNet to further validate the effectiveness and robustness of our approach.
Table~\ref{improvements} shows the performance comparison on four benchmark datasets, \ie, ADE20K, Cityscapes, LIP and COCO-Stuff.
For a fair comparison, all the results are obtained under the single-scale testing without using the proposed coarse-to-fine iterative inference strategy.
It is observed that MCIBI++ consistently improves the segmentation performance of the four baseline frameworks by large margins on all datasets.
For example, for the segmentation framework without within image context module $\mathcal{A}_{wi}$ (\ie, FCN), MCIBI++ brings $6.43\%$ mIoU improvements on ADE20K, $3.61\%$ mIoU improvements on Cityscapes, $2.50\%$ mIoU improvements on LIP and $5.62\%$ mIoU improvements on COCO-Stuff, respectively.
For a stronger baseline with $\mathcal{A}_{wi}$ (\eg, DeepLabV3), MCIBI++ boosts the mIoU performance from $43.19\%$ to $44.85\%$ on ADE20K, $79.62\%$ to $80.72\%$ on Cityscapes, $52.35\%$ to $53.59\%$ on LIP and $37.63\%$ to $38.94\%$ on COCO-Stuff, respectively.
These comparison demonstrates the flexibility and generalizability of our proposed MCIBI++ paradigm.

\subsection{Comparison with the State-of-the-Art Methods}

\para{Results on ADE20K.}
Table~\ref{sotaade20k} compares the segmentation results on the validation set of ADE20K.
With the ResNet-101 backbone, ISNet~\cite{jin2021isnet} achieves $47.31\%$ mIoU via integrating the image-level and semantic-level context for semantic segmentation, which is the previous best method.
Based on the same ResNet-101 backbone, our approach achieves superior mIoU of $47.93\%$ with a significant margin over ISNet.
Moreover, our UperNet+MCIBI++ with the ResNet-101 backbone even outperforms the previous ISNet, DeepLabV3 and DeepLabV3+MCIBI with a more powerful ResNeSt-101~\cite{zhang2020resnest} backbone by $0.38\%$, $1.02\%$ and $0.57\%$, respectively.
Without transformer-based backbone, the state-of-the-art segmentation framework is DeepLabV3 with ResNeSt-200, hitting $48.36\%$ mIoU.
Due to the superiority of the proposed MCIBI++, our UperNet+MCIBI++ with a relatively light backbone network ResNeSt-101 still reports the new state-of-the-art performance (\ie, $48.56\%$ mIoU) among the transformer-free methods.
In addition, prior to this work, UperNet achieves the state-of-the-art with $53.50\%$ mIoU by adopting a transformer-based backbone network, \ie, Swin-Large \cite{liu2021swin}.
Based on this, our method UperNet+MCIBI++ with the same Swin-Large backbone achieves a new state-of-the-art result with mIoU hitting $54.52\%$.
It is worth noting that ADE20K is very challenging due to its various image sizes, the large number of semantic categories and the gap between the training and validation sets.
Our framework achieves consistent improvements over counterparts and new state-of-the-art performance on this dataset, which clearly demonstrates the significance of our proposed MCIBI++ paradigm.

\begin{table}[t]
\centering
\caption{
   Segmentation results on the validation set of LIP.
   Test time augmentation is utilized for fair comparison.
}\label{sotalip}
\vspace{-0.30cm}
\resizebox{.48\textwidth}{!}{
\begin{tabular}{c|c|c|c}
   \hline
   \hline
   Method                                           &Backbone      &Stride       &mIoU ($\%$)     \\
   \hline
   Attention \cite{chen2016attention}               &ResNet-101    &-            &42.92           \\
   DeepLab \cite{chen2017deeplab}                   &ResNet-101    &-            &44.80           \\
   MMAN \cite{luo2018macro}                         &ResNet-101    &-            &46.81           \\
   CE2P \cite{ruan2019devil}                        &ResNet-101    &$16\times$   &53.10           \\
   BraidNet \cite{liu2019braidnet}                  &ResNet-101    &$8\times$    &54.54           \\
   DeepLabV3 \cite{chen2017rethinking}              &ResNet-101    &$8\times$    &54.58           \\
   OCNet \cite{yuan2018ocnet}                       &ResNet-101    &$8\times$    &54.72           \\
   CCNet \cite{huang2019ccnet}                      &ResNet-101    &$8\times$    &55.47           \\
   OCRNet \cite{yuan2019object}                     &ResNet-101    &$8\times$    &55.60           \\
   CorrPM \cite{zhang2021correlation}               &ResNet-101    &$8\times$    &55.33           \\
   ISNet \cite{jin2021isnet}                        &ResNet-101    &$8\times$    &55.41           \\
   ISNet \cite{jin2021isnet}                        &ResNeSt-101   &$8\times$    &56.81           \\
   HRNet \cite{wang2020deep}                        &HRNetV2-W48   &$4\times$    &55.90           \\
   OCRNet \cite{yuan2019object}                     &HRNetV2-W48   &$4\times$    &56.65           \\
   \hline
   DeepLabV3+MCIBI \cite{jin2021mining}             &ResNet-101    &$8\times$    &55.42           \\
   DeepLabV3+MCIBI \cite{jin2021mining}             &ResNeSt-101   &$8\times$    &56.34           \\
   DeepLabV3+MCIBI \cite{jin2021mining}             &HRNetV2-W48   &$4\times$    &56.99           \\
   \hline
   UperNet+MCIBI++ (\emph{ours})                    &ResNet-101    &$8\times$    &56.32           \\
   UperNet+MCIBI++ (\emph{ours})                    &ResNeSt-101   &$8\times$    &57.08           \\
   UperNet+MCIBI++ (\emph{ours})                    &Swin-Large    &$32\times$   &\textbf{59.91}  \\
   \hline
   \hline
\end{tabular}}
\vspace{-0.30cm}
\end{table}

\para{Results on LIP.}
Table~\ref{sotalip} shows the comparison results on the validation set of LIP.
If only considering the ResNet-101 backbone, we observe that UperNet+MCIBI++ yields an mIoU of $56.32\%$, which is $0.72\%$ higher than the previous best segmentation framework OCRNet.
Meanwhile, it can be observed that our UperNet+MCIBI++ with the ResNeSt-101 backbone achieves $57.08\%$ mIoU, on par with state-of-the-art approaches.
To further show the effectiveness of our method, we adopt UperNet+MCIBI++ with the transformer-based network Swin-Large as backbone and report the new state-of-the-art performance, hitting $59.91\%$ mIoU.
This is particularly impressive considering that human parsing is a more challenging fine-grained semantic segmentation task.

\begin{figure*}
\centering
\includegraphics[width=1.0\textwidth]{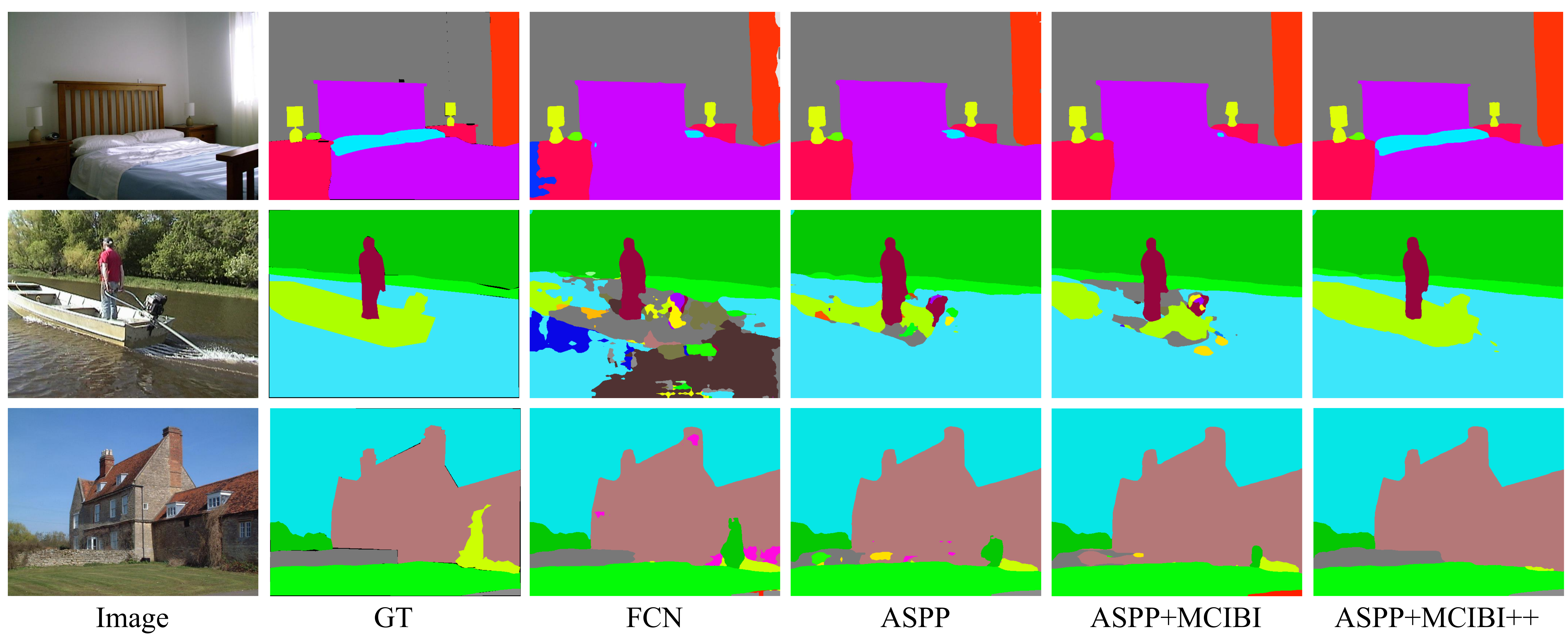}
\vspace{-0.70cm}
\caption{
   Visual comparisons between Ground Truth (GT), FCN, ASPP, ASPP+MCIBI and ASPP+MCIBI++(\emph{ours}) on the validation set of ADE20K.
}\label{compare_ade20k}
\end{figure*}

\para{Results on COCO-Stuff.}
Table~\ref{sotacocostuff} reports the performance comparison of our method against nine competitors on the test set of COCO-Stuff.
In the case of employing ResNet-101 backbone network, UperNet+MCIBI++ reaches $41.84\%$ mIoU, outperforming the state-of-the-art ISNet by $0.24\%$ mIoU.
It is noteworthy that this performance is also superior to some methods with a stronger backbone, \eg, OCRNet with HRNetV2-W48 ($41.84\%$ \emph{v.s.} $40.50\%$).
With a more robust feature extractor ResNeSt-101, our method achieves mIoU of $42.71\%$, which surpasses the previous CNN-based best result (\ie, DeepLabV3+MCIBI) by $0.56\%$ mIoU.
Since recent academic attention has been geared toward transformer-based backbone networks, in Table~\ref{sotacocostuff}, we have also integrated our method into the popular Swin-Large backbone.
As we can see, UperNet+MCIBI++ with Swin-Large obtains $50.27\%$ mIoU, which outperforms other competitors by a large margin.
These results firmly suggest the superiority of the proposed soft mining the contextual information beyond the input image paradigm.

\begin{table}[t]
\centering
\caption{
   Comparison with the state-of-the-art segmentation frameworks on the test set of COCO-Stuff.
   Multi-scale and flipping testing are employed for fair comparison.
   The best score is marked in \textbf{bold}.
}\label{sotacocostuff}
\vspace{-0.3cm}
\resizebox{.48\textwidth}{!}{
\begin{tabular}{c|c|c|c}
   \hline
   \hline
   Method                                           &Backbone      &Stride       &mIoU ($\%$)       \\
   \hline
   DANet \cite{fu2019dual}                          &ResNet-101    &$8\times$    &39.70             \\
   OCRNet \cite{yuan2019object}                     &ResNet-101    &$8\times$    &39.50             \\
   SVCNet \cite{ding2019semantic}                   &ResNet-101    &$8\times$    &39.60             \\
   EMANet \cite{li2019expectation}                  &ResNet-101    &$8\times$    &39.90             \\
   ACNet \cite{fu2019adaptive}                      &ResNet-101    &$8\times$    &40.10             \\
   ISNet \cite{jin2021isnet}                        &ResNet-101    &$8\times$    &41.60             \\
   MaskFormer \cite{cheng2021per}                   &ResNet-101    &$8\times$    &39.30             \\
   ISNet \cite{jin2021isnet}                        &ResNeSt-101   &$8\times$    &42.08             \\
   OCRNet \cite{yuan2019object}                     &HRNetV2-W48   &$4\times$    &40.50             \\
   \hline
   DeepLabV3+MCIBI \cite{jin2021mining}             &ResNet-101    &$8\times$    &41.49             \\
   DeepLabV3+MCIBI \cite{jin2021mining}             &ResNeSt-101   &$8\times$    &42.15             \\
   DeepLabV3+MCIBI \cite{jin2021mining}             &ViT-Large     &$16\times$   &44.89             \\
   \hline
   UperNet+MCIBI++ (\emph{ours})                    &ResNet-101    &$8\times$    &41.84             \\
   UperNet+MCIBI++ (\emph{ours})                    &ResNeSt-101   &$8\times$    &42.71             \\
   UperNet+MCIBI++ (\emph{ours})                    &Swin-Large    &$32\times$   &\textbf{50.27}    \\
   \hline
   \hline
\end{tabular}}
\end{table}

\begin{figure}
\centering
\includegraphics[width=0.45\textwidth]{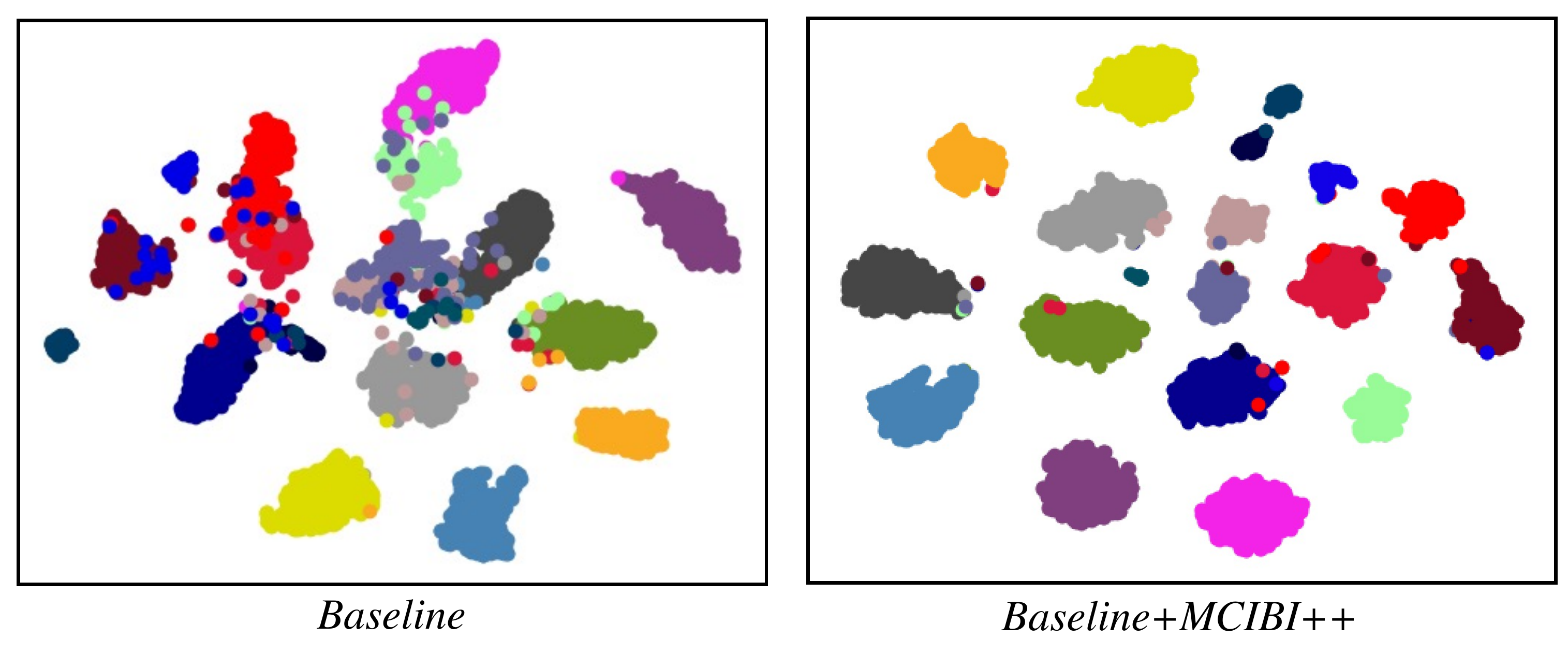}
\vspace{-0.3cm}
\caption{
   The t-SNE visualization on Cityscapes validation set.
   We first compute the composite vector of the pixel representations of the same category within an image and then denote it as a point.
   The official recommended color table is leveraged to color these points according to their category labels.
   Obviously, soft mining contextual information beyond image (MCIBI++) can beget a more well-structured semantic feature space.
}\label{tsne}
\vspace{-0.3cm}
\end{figure}

\begin{table}[t]
\centering
\caption{
   Comparison of performance on the test set of Cityscapes with state-of-the-art approaches (trained on trainval set).
   Multi-scale and flipping testing are used for fair comparison.
   \textbf{Bold} denotes the best score.
}\label{sotacityscapes}
\vspace{-0.3cm}
\resizebox{.48\textwidth}{!}{
\begin{tabular}{c|c|c|c}
   \hline
   \hline
   Method                                           &Backbone      &Stride       &mIoU ($\%$) \\
   \hline
   CCNet \cite{huang2019ccnet}                      &ResNet-101    &$8\times$    &81.90     \\
   PSPNet \cite{zhao2017pyramid}                    &ResNet-101    &$8\times$    &78.40     \\
   PSANet \cite{zhao2018psanet}                     &ResNet-101    &$8\times$    &80.10     \\
   OCNet \cite{yuan2018ocnet}                       &ResNet-101    &$8\times$    &80.10     \\
   OCRNet \cite{yuan2019object}                     &ResNet-101    &$8\times$    &81.80     \\
   DANet \cite{fu2019dual}                          &ResNet-101    &$8\times$    &81.50     \\
   ACFNet \cite{zhang2019acfnet}                    &ResNet-101    &$8\times$    &81.80     \\
   ANNet \cite{zhu2019asymmetric}                   &ResNet-101    &$8\times$    &81.30     \\
   ACNet \cite{fu2019adaptive}                      &ResNet-101    &$8\times$    &82.30     \\
   CPNet \cite{yu2020context}                       &ResNet-101    &$8\times$    &81.30     \\
   SFNet \cite{li2020semantic}                      &ResNet-101    &$8\times$    &81.80     \\
   SPNet \cite{hou2020strip}                        &ResNet-101    &$8\times$    &82.00     \\
   DenseASPP \cite{yang2018denseaspp}               &DenseNet-161  &$8\times$    &80.60     \\
   Dynamic \cite{li2020learning}                    &Layer33-PSP   &$8\times$    &80.70     \\
   HRNet \cite{wang2020deep}                        &HRNetV2-W48   &$4\times$    &81.60     \\
   OCRNet \cite{yuan2019object}                     &HRNetV2-W48   &$4\times$    &82.40     \\
   \hline
   DeepLabV3+MCIBI \cite{jin2021mining}             &ResNet-101    &$8\times$    &82.03     \\
   DeepLabV3+MCIBI \cite{jin2021mining}             &ResNeSt-101   &$8\times$    &81.59     \\
   DeepLabV3+MCIBI \cite{jin2021mining}             &HRNetV2-W48   &$4\times$    &82.55     \\
   \hline
   DeepLabV3+MCIBI++ (\emph{ours})                  &ResNet-101    &$8\times$    &82.20     \\
   DeepLabV3+MCIBI++ (\emph{ours})                  &ResNeSt-101   &$8\times$    &81.70      \\
   DeepLabV3+MCIBI++ (\emph{ours})                  &HRNetV2-W48   &$4\times$    &\textbf{82.74}     \\
   \hline
   \hline
\end{tabular}}
\vspace{-0.3cm}
\end{table}

\para{Results on Cityscapes.}
Table~\ref{sotacityscapes} shows the comparison results on the test set of Cityscapes.
For a fair comparison, only the frameworks trained on the Cityscapes \emph{trainval set} are added in this table.
As observed, we first utilize the common ResNet-101 backbone network to train MCIBI++ and obtain $82.20\%$ mIoU, 
which surpasses the previous best method with the same backbone by $0.17\%$ mIoU ($82.20\%$ \emph{v.s.} $82.03\%$).
Note that, $82.03\%$ mIoU for Cityscapes under the comparable settings (\eg, the adopted pre-trained dataset, backbone, \etc) is a strong baseline.
Moreover, we can observe that DeepLabV3+MCIBI++ with HRNetV2-W48 achieves $82.74\%$ mIoU, which is $0.19\%$ higher than the previous best result $82.55\%$ mIoU obtained by DeepLabV3+MCIBI.
These experimental results further manifest the effectiveness of our approach.

\begin{table}[t]
\centering
\caption{
   Quantitative evaluation on the PASCAL-Context validation set. 
   Multi-scale and flipping testing are used for fair comparison.
}\label{sotapascalcontext}
\vspace{-0.3cm}
\resizebox{.48\textwidth}{!}{
\begin{tabular}{c|c|c|c}
   \hline
   \hline
   Method                                           &Backbone      &Stride       &mIoU ($\%$)        \\
   \hline
   EncNet \cite{zhang2018context}                   &ResNet-101    &$8\times$    &51.70              \\
   DANet \cite{fu2019dual}                          &ResNet-101    &$8\times$    &52.60              \\
   ANNet \cite{zhu2019asymmetric}                   &ResNet-101    &$8\times$    &52.80              \\
   SVCNet \cite{ding2019semantic}                   &ResNet-101    &$8\times$    &53.20              \\
   CPNet \cite{yu2020context}                       &ResNet-101    &$8\times$    &53.90              \\
   APCNet \cite{he2019adaptive}                     &ResNet-101    &$8\times$    &54.70              \\
   ACNet \cite{fu2019adaptive}                      &ResNet-101    &$8\times$    &54.10              \\
   EMANet \cite{li2019expectation}                  &ResNet-101    &$8\times$    &53.10              \\
   SFNet \cite{li2020semantic}                      &ResNet-101    &$8\times$    &53.80              \\
   OCRNet \cite{yuan2019object}                     &ResNet-101    &$8\times$    &54.80              \\
   SPNet \cite{hou2020strip}                        &ResNet-101    &$8\times$    &54.50              \\
   HRNet \cite{wang2020deep}                        &HRNetV2-W48   &$4\times$    &54.00              \\
   OCRNet \cite{yuan2019object}                     &HRNetV2-W48   &$4\times$    &56.20              \\
   \hline
   UperNet+MCIBI++ (\emph{ours})                    &ResNet-101    &$8\times$    &56.82              \\
   UperNet+MCIBI++ (\emph{ours})                    &ResNeSt-101   &$8\times$    &57.92              \\
   UperNet+MCIBI++ (\emph{ours})                    &Swin-Large    &$32\times$   &\textbf{64.01}     \\
   \hline
   \hline
\end{tabular}}
\end{table}

\begin{table}[t]
\centering
\caption{
   The semantic segmentation result on the validation set of PASCAL VOC 2012.
   Single-scale testing is utilized for fair comparison.
}\label{sotavocaug}
\vspace{-0.3cm}
\resizebox{.48\textwidth}{!}{
\begin{tabular}{c|c|c|c}
   \hline
   \hline
   Method                                                       &Backbone      &Stride       &mIoU ($\%$)        \\
   \hline
   FCN \cite{long2015fully} (\emph{our impl.})                  &ResNet-101    &$8\times$    &70.59              \\
   SemanticFPN \cite{Kirillov_2019} (\emph{our impl.})          &ResNet-101    &$8\times$    &72.51              \\
   PointRend \cite{kirillov2020pointrend} (\emph{our impl.})    &ResNet-101    &$8\times$    &72.31              \\
   ANNet \cite{zhu2019asymmetric} (\emph{our impl.})            &ResNet-101    &$8\times$    &78.15              \\
   APCNet \cite{he2019adaptive} (\emph{our impl.})              &ResNet-101    &$8\times$    &78.99              \\
   OCRNet \cite{yuan2019object} (\emph{our impl.})              &ResNet-101    &$8\times$    &78.82              \\
   CCNet \cite{huang2019ccnet} (\emph{our impl.})               &ResNet-101    &$8\times$    &78.02              \\
   CE2P \cite{ruan2019devil} (\emph{our impl.})                 &ResNet-101    &$8\times$    &77.77              \\
   DANet \cite{fu2019dual} (\emph{our impl.})                   &ResNet-101    &$8\times$    &77.97              \\
   DeepLabV3 \cite{chen2017rethinking} (\emph{our impl.})       &ResNet-101    &$8\times$    &79.52              \\
   DeepLabV3Plus \cite{chen2018encoder} (\emph{our impl.})      &ResNet-101    &$8\times$    &79.19              \\
   DMNet \cite{he2019dynamic} (\emph{our impl.})                &ResNet-101    &$8\times$    &79.15              \\
   ISANet \cite{huang2019interlaced} (\emph{our impl.})         &ResNet-101    &$8\times$    &78.60              \\
   UperNet \cite{xiao2018unified} (\emph{our impl.})            &ResNet-101    &$8\times$    &79.13              \\
   PSPNet \cite{zhao2017pyramid} (\emph{our impl.})             &ResNet-101    &$8\times$    &79.04              \\
   PSANet \cite{zhao2018psanet} (\emph{our impl.})              &ResNet-101    &$8\times$    &78.97              \\
   Nonlocal \cite{wang2018non} (\emph{our impl.})               &ResNet-101    &$8\times$    &78.89              \\
   GCNet \cite{cao2019gcnet} (\emph{our impl.})                 &ResNet-101    &$8\times$    &78.81              \\
   EncNet \cite{zhang2018context} (\emph{our impl.})            &ResNet-101    &$8\times$    &77.61              \\
   EMANet \cite{li2019expectation} (\emph{our impl.})           &ResNet-101    &$8\times$    &76.43              \\
   DNLNet \cite{yin2020disentangled} (\emph{our impl.})         &ResNet-101    &$8\times$    &78.37              \\
   OCRNet \cite{yuan2019object} (\emph{our impl.})              &HRNetV2-W48   &$4\times$    &77.60              \\
   \hline
   UperNet+MCIBI++ (\emph{ours})                                &ResNet-50     &$8\times$    &79.48              \\
   UperNet+MCIBI++ (\emph{ours})                                &ResNet-101    &$8\times$    &\textbf{80.42}     \\
   \hline
   \hline
\end{tabular}}
\vspace{-0.3cm}
\end{table}

\begin{figure}
\centering
\includegraphics[width=0.48\textwidth]{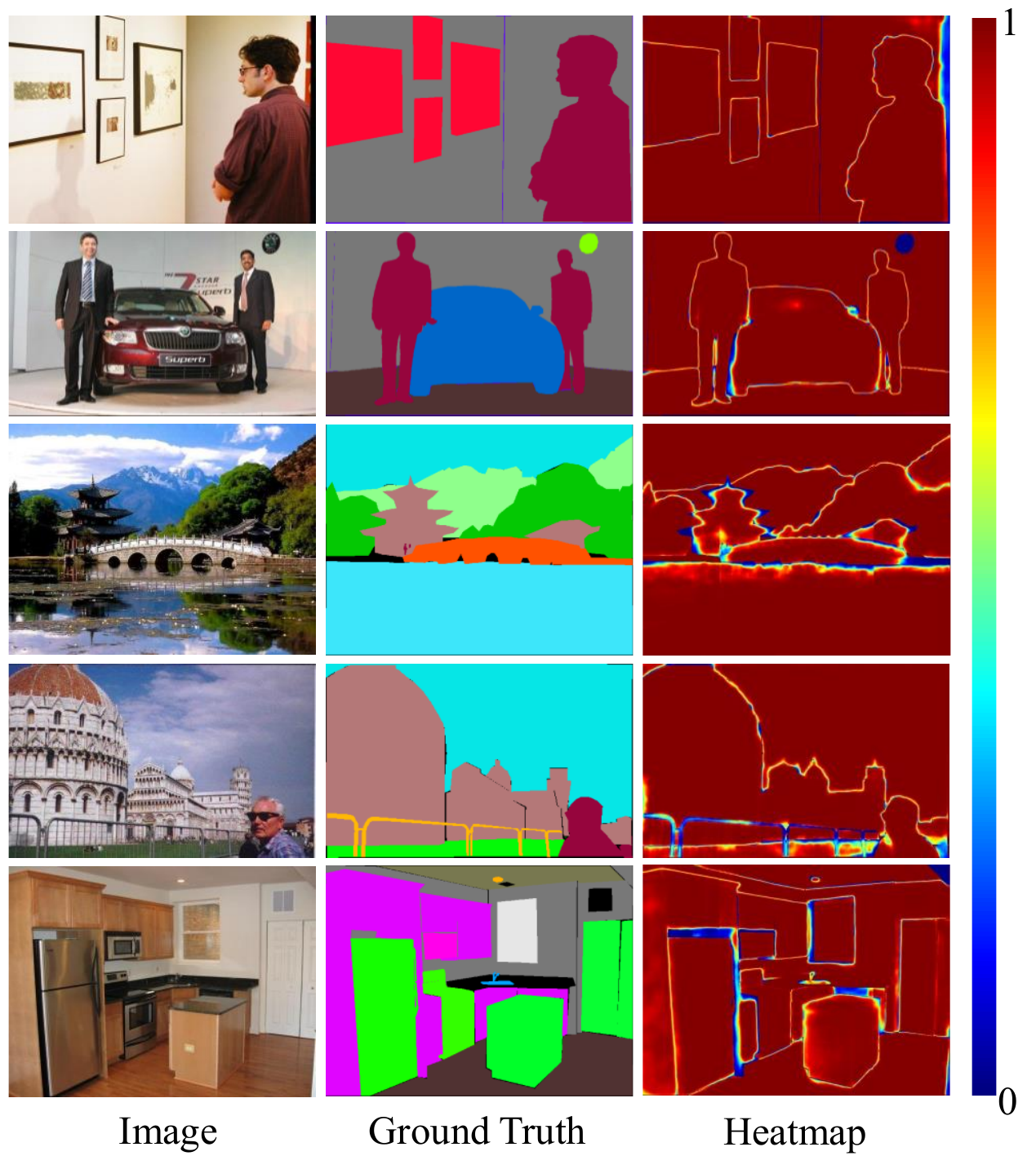}
\caption{
   Visualization of the proposed dataset-level context aggregation scheme. 
   The pixel value of the heatmap shows the aggregation weight of the corresponding ground truth category.
}\label{dca}
\vspace{-0.30cm}
\end{figure}

\begin{figure*}
\centering
\includegraphics[width=1.0\textwidth]{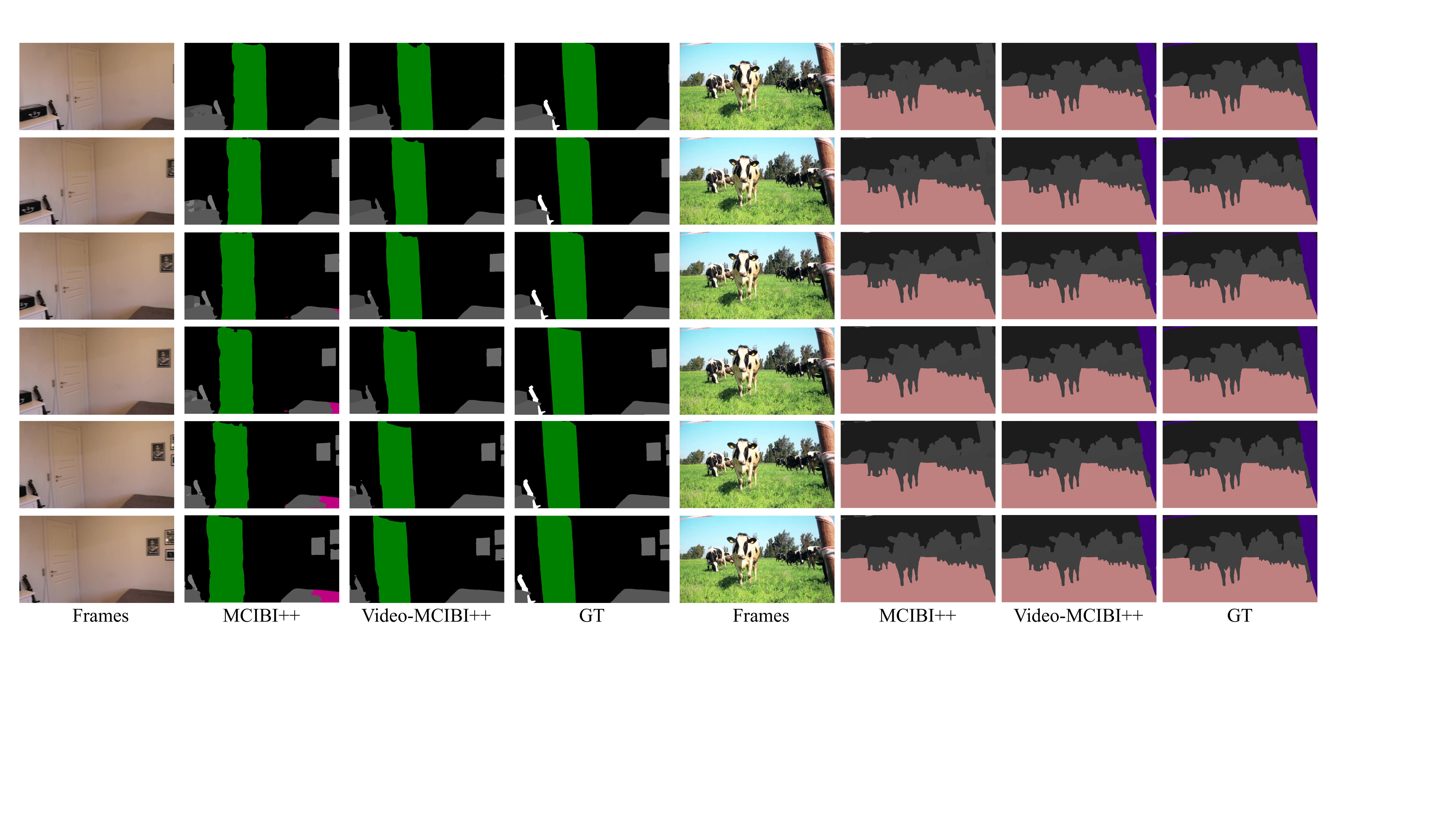}
\caption{
   Qualitative comparisons of the segmentation results on VSPW validation set.
}\label{compare_vspw}
\vspace{-0.40cm}
\end{figure*}

\subsection{Additional Results on Other Benchmarks}
Beyond the four challenging image semantic segmentation benchmarks above, we also test MCIBI++ on other benchmarks, including PASCAL-Context \cite{everingham2010pascal} and PASCAL VOC 2012~\cite{everingham2015pascal, hariharan2011semantic}.

\para{Results on PASCAL-Context.}
PASCAL-Context is a challenging scene parsing dataset that provides $4,998$ images for training and $5,105$ images for validation.
We trained the segmentation frameworks for $260$ epochs with batch size of $16$, crop size of $480 \times 480$ with initial learning rate of $0.004$.
For evaluation, we performed multi-scale testing with the horizontal flipping operation.
As illustrated in Table~\ref{sotapascalcontext}, we can see that UperNet+MCIBI++ with ResNet-101 as the backbone reaches $56.82\%$ mIoU, achieving the new state-of-the-art performance among the ResNet-101-based approaches.
This result is also $0.62\%$ mIoU higher than OCRNet with a more robust backbone HRNetV2-W48, which is the previous best segmentor on PASCAL-Context.
In addition, we also train UperNet+MCIBI++ with stronger backbone networks, \ie, ResNeSt-101 and Swin-Large and achieve $57.92\%$ and $64.01\%$ mIoU respectively.
These results outperform the previous best approach by large margins, \ie, $1.72\%$ mIoU and $7.81\%$ mIoU, respectively.
These improvements are nontrivial, given that OCRNet with HRNetV2-W48 is a strong segmentor and the PASCAL-Context task is fairly saturated.

\para{Results on PASCAL VOC 2012.}
We also evaluate MCIBI++ on the PASCAL VOC 2012 semantic segmentation benchmark, which contains $10, 582$ (\emph{trainaug}) training images, $1, 449$ validation images and $1, 456$ testing images.
Table~\ref{sotavocaug} shows the comparison results.
We report our PyTorch re-implementation results so that all the reported methods can be trained and evaluated on the same environment and settings for a fair comparison. 
The DeepLabV3 with the ResNet-101 backbone (\ie, $79.52\%$ mIoU) is the previous best algorithm.
Our UperNet+MCIBI++ with the same backbone network ResNet-101 obtains a superior mIoU of $80.42\%$ with a significant margin over DeepLabV3, demonstrating the effectiveness of our method.

\subsection{Qualitative Results}

\para{Visual Comparison.}
The visual comparisons among FCN, ASPP, ASPP+MCIBI and ASPP+MCIBI++ are shown in Fig.~\ref{compare_ade20k}.
As we can see, ASPP+MCIBI++ achieves better visual results than others. 
For example, for the first case, only our segmentation framework can extract the pillows from the bed accurately.
And for the second case, our algorithm can correctly segment the boat while the other approaches fail to achieve this point.

\para{Visualization of Learned Features.} 
As illustrated in Fig.~\ref{tsne}, we visualize the pixel representations learned by the baseline model (left) and our framework (right), respectively.
Specifically, we first calculate the composite feature vectors of the pixel representations of the same category within an image.
Then, we employ t-SNE algorithm to project these feature vectors into a two-dimensional plane.
As we can see, after incorporating the dataset-level information, the features of different categories are more discriminative, \ie, the pixel representations with the same category are more compact and the features of various classes are well separated.
This comparison shows that the dataset-level category representations can help augment the pixel representations by improving the spatial distribution of the original features in the non-linear embedding space so that the network can classify the pixel more accurately.

\para{Visualization of Dataset-Level Context Aggregation.}
Fig.~\ref{dca} exhibits the visualization results of the proposed dataset-level context aggregation (DCA) scheme on the samples from ADE20K datasets.
We can observe that most of the pixels can accurately aggregate the demanded dataset-level category representations, demonstrating the effectiveness of the proposed dataset-level context aggregation scheme.

\begin{table}[t]
\centering
\caption{
   Ablation study of Video-MCIBI++ on VSPW validation set.
   The adopted backbone is ResNet-50.
   $R_{N-i}$ denotes that $C_{bi}^{N-i}$ in Eq.~(\ref{eq19}) is replaced with $R_{N-i}$.
}\label{vswpablation}
\vspace{-0.3cm}
\resizebox{.48\textwidth}{!}{
\begin{tabular}{c|c|cc|c}
   \hline
   \hline
   Method                      &Frames       &$R_{N-i}$         &$C_{bi}^{N-i}$      &mIoU ($\%$)      \\
   \hline
   UperNet+MCIBI++             &-            &-                 &-                   &41.08            \\
   \hline
   UperNet+Video-MCIBI++       &1            &\checkmark        &                    &41.54            \\
   UperNet+Video-MCIBI++       &1            &                  &\checkmark          &42.06            \\
   \hline
   UperNet+Video-MCIBI++       &2            &\checkmark        &                    &41.85            \\
   UperNet+Video-MCIBI++       &2            &                  &\checkmark          &42.39            \\
   \hline
   UperNet+Video-MCIBI++       &3            &\checkmark        &                    &41.07            \\
   UperNet+Video-MCIBI++       &3            &                  &\checkmark          &42.40            \\
   \hline
   \hline
\end{tabular}}
\vspace{-0.3cm}
\end{table}

\begin{table*}[t]
\centering
\caption{
   Comparison on the VSPW validation set. mVC$_C$ means we use a clip with $C$ frames.
   MCIBI and MCIBI++ represent for the image semantic segmentation methods proposed in ths paper while
   Video-MCIBI and Video-MCIBI++ denote the video semantic segmentation approaches.
}\label{sotavspw}
\vspace{-0.3cm}
\resizebox{1.0\textwidth}{!}{
\begin{tabular}{cc|ccc|ccc}
   \hline
   \hline
   Method                                                 &Backbone       &Pixel Accuracy ($\%$)                &Mean Class Accuracy ($\%$)           &mIoU ($\%$)                          &TC ($\%$)                            &mVC$_8$ ($\%$)                       &mVC$_{16}$ ($\%$)                    \\
   \hline
   \textbf{\emph{Image Semantic Segmentation Framework}} &&&&&&&\\
   DeepLabV3Plus \cite{miao2021vspw}                      &ResNet-101     &-                                    &-                                    &34.67                                &65.45                                &83.24                                &78.24                                \\
   UperNet \cite{miao2021vspw}                            &ResNet-101     &-                                    &-                                    &36.46                                &63.10                                &82.55                                &76.08                                \\
   PSPNet \cite{miao2021vspw}                             &ResNet-101     &-                                    &-                                    &36.47                                &65.89                                &84.16                                &79.63                                \\
   OCRNet \cite{miao2021vspw}                             &ResNet-101     &-                                    &-                                    &36.68                                &66.21                                &83.97                                &79.04                                \\
   UperNet+MCIBI \cite{jin2021mining,jin2021memory}       &ResNet-101     &73.85                                &53.42                                &42.11                                &68.15                                &82.11                                &76.52                                \\
   UperNet+MCIBI \cite{jin2021mining,jin2021memory}       &Swin-Large     &81.22                                &64.65                                &55.18                                &75.75                                &89.52                                &86.13                                \\
   UperNet+MCIBI++ (\emph{ours})                          &ResNet-101     &75.15 (\textbf{{\color{teal}+1.30}}) &55.64 (\textbf{{\color{teal}+2.22}}) &43.21 (\textbf{{\color{teal}+1.10}}) &70.06 (\textbf{{\color{teal}+1.91}}) &83.62 (\textbf{{\color{teal}+1.51}}) &78.48 (\textbf{{\color{teal}+1.96}}) \\
   UperNet+MCIBI++ (\emph{ours})                          &Swin-Large     &81.70 (\textbf{{\color{teal}+0.48}}) &65.43 (\textbf{{\color{teal}+0.78}}) &56.04 (\textbf{{\color{teal}+0.86}}) &76.40 (\textbf{{\color{teal}+0.65}}) &90.07 (\textbf{{\color{teal}+0.55}}) &87.01 (\textbf{{\color{teal}+0.88}}) \\
   \hline
   \textbf{\emph{Video Semantic Segmentation Framework}} &&&&&&&\\
   ETC+PSPNet \cite{miao2021vspw}                         &ResNet-101     &-                                    &-                                    &36.55                                &67.94                                &84.10                                &79.22                                \\
   NetWarp+PSPNet \cite{miao2021vspw}                     &ResNet-101     &-                                    &-                                    &36.95                                &67.85                                &84.36                                &79.42                                \\
   ETC+OCRNet \cite{miao2021vspw}                         &ResNet-101     &-                                    &-                                    &37.46                                &68.99                                &84.10                                &79.10                                \\
   NetWarp+OCRNet \cite{miao2021vspw}                     &ResNet-101     &-                                    &-                                    &37.52                                &68.89                                &84.00                                &78.97                                \\
   TCB$_{\emph{st-ppm}}$ \cite{miao2021vspw}              &ResNet-101     &-                                    &-                                    &37.46                                &70.30                                &86.95                                &82.12                                \\
   TCB$_{\emph{st-ocr}}$ \cite{miao2021vspw}              &ResNet-101     &-                                    &-                                    &37.40                                &72.20                                &86.88                                &82.04                                \\
   TCB$_{\emph{st-ocr~mem}}$ \cite{miao2021vspw}          &ResNet-101     &-                                    &-                                    &37.82                                &73.63                                &87.86                                &83.99                                \\
   UperNet+Video-MCIBI \cite{jin2021mining,jin2021memory} &Swin-Large     &81.97                                &68.58                                &57.57                                &80.90                                &90.67                                &87.39                                \\
   UperNet+Video-MCIBI++ (\emph{ours})                    &Swin-Large     &82.20 (\textbf{{\color{teal}+0.23}}) &69.00 (\textbf{{\color{teal}+0.42}}) &58.88 (\textbf{{\color{teal}+1.31}}) &81.21 (\textbf{{\color{teal}+0.31}}) &91.53 (\textbf{{\color{teal}+0.86}}) &88.50 (\textbf{{\color{teal}+1.11}}) \\
   \hline
   \hline
\end{tabular}}
\vspace{-0.3cm}
\end{table*}

\section{Experiments on Video Semantic Segmentation}
In this section, we show the results of our proposed video semantic segmentation paradigm in Section \ref{method:VSS} (named as Video-MCIBI++ for distinction), where the memory module is also used to integrate the temporal information.
Specifically, the proposed Video-MCIBI++ paradigm leverages the memory module to incorporate both the temporal information (\emph{i.e.}, the relations to the previous frames) and the dataset-level contextual information (\emph{i.e.}, the relations to the frames from all trained videos) to the segmentation frameworks.

\subsection{Experimental Setup}

\para{Dataset.}
We validate our framework on the recent released VSPW dataset~\cite{miao2021vspw}.
There are $3, 536$ annotated videos, including $251, 633$ frames from $124$ categories.
The train, validation and test sets of VSPW contain $2, 806$/$343$/$387$ videos with $198,244$/$24,502$/$28,887$ frames, respectively.
It is a quite challenging video semantic segmentation benchmark due to its scale, diversity, long-temporal and high frame rate.

\para{Implementation Details.}
We trained our segmentors on eight NVIDIA Tesla V100 GPUs with a 32 GB memory per-card.
The backbone network was initialized with the weights pre-trained on the ImageNet~\cite{krizhevsky2012imagenet} and the newly added modules were randomly initialized.
Following~\cite{miao2021vspw}, all frames in VSPW were resized to keep their shorter side being $512$.
The optimizer was stochastic gradient descent (SGD) with momentum of $0.9$, weight decay of $0.0005$, batch size of $16$, and initial learning rate of $0.01$.
By default, the segmentation frameworks were trained for $42,000$ iterations.
We adopted random horizontal flipping, random scaling and color jitter as the data augmentation.
All other hyper-parameters followed the settings in the codebase of VSPW~\cite{miao2021vspw}, if not specified.

\para{Evaluation Metrics.}
Following previous studies, pixel accuracy, mean class accuracy, mean intersection-over-union (mIoU), temporal consistency (TC) and video consistency (VC) are employed for evaluation.
Please refer to~\cite{miao2021vspw} for more details about the calculation method of TC and VC.

\subsection{Ablation Studies}

\para{Number of frames.}
Table \ref{vswpablation} shows the ablation studies on the number of the used historical frames in Eq.~(\ref{eq18})-(\ref{eq19}).
It is observed that the mIoU improvements start to be limited when the number of the historical frames increases to three. 
Thus, we utilize two historical frames to incorporate the temporal information in Video-MCIBI++ by default.

\para{Compare $C_{bi}^{N-i}$ with $R_{N-i}$.}
As discussed in Section \ref{method:VSS}, we think that utilizing $C_{bi}^{N-i}$ in Eq.~(\ref{eq19}) is better than adopting $R_{N-i}$.
We also conduct the ablation experiments in Table \ref{vswpablation} to validate our points.
From the results, it is observed that using $C_{bi}^{N-i}$ in our framework outperforms the counterpart using $R_{N-i}$ by $0.52\%$, $0.54\%$ and $1.33\%$, respectively, when the number of the historical frames increases from 1 to 3.
These results indicate that introducing $C_{bi}^{N-i}$ can help the network capture the temporal relations more effectively.
In addition, it is observed that the performance of $R_{N-i}$ decreases by $0.68\%$ when the number of the historical frames increases from 2 to 3.
This drop indicates that $R_{N-i}$ may suppress the segmentation performance when its proportion in $R_{aug}$ grows to a certain level.

\para{Compare Video-MCIBI++ with MCIBI++.}
We first show the results of image semantic segmentation frameworks (\ie, UperNet+MCIBI and UperNet+MCIBI++) on video semantic segmentation task (\ie, taking each frame as an input image) in Table \ref{sotavspw}.
From these results, we can observe that UperNet+MCIBI++ with the ResNet-101 backbone achieves $43.21\%$ mIoU, $70.06\%$ TC and $83.62\%$ mVC$_{8}$ score on the validation set of VSPW, outperforming UperNet with the same backbone by $6.74\%$ mIoU, $6.96\%$ TC and $1.07\%$ mVC$_{8}$ score.
The comparison further demonstrates the effectiveness of MCIBI++ and shows that UperNet+MCIBI++ is an efficient baseline framework for video semantic segmentation task.
Also, it is observed that UperNet+Video-MCIBI++ with the Swin-Large backbone reaches $58.88\%$ mIoU, $81.21\%$ TC and $91.53\%$ mVC$_{8}$, which is $2.84\%$ mIoU, $4.81\%$ TC and $1.46\%$ mVC$_{8}$ higher than the baseline method, \ie, UperNet+MCIBI++ with Swin-Large.
This result indicates that extracting the dataset-level category representations of the historical frames can well supplement the temporal information in the segmentation framework, which helps boost the segmentation performance and stability.
In addition, UperNet+Video-MCIBI with Swin-Large also outperforms UperNet+MCIBI with Swin-Large by $2.39\%$ mIoU, $5.15\%$ TC and $1.15\%$ mVC$_{8}$, which is consistent with the comparison result between UperNet+MCIBI++ and UperNet+Video-MCIBI++.
The consistent performance improvement further validates the effectiveness of our proposed video semantic segmentation framework.

\para{The comparison between Video-MCIBI++ and Video-MCIBI.}
Table \ref{sotavspw} also compares the segmentation performance of Video-MCIBI++ and Video-MCIBI.
The key difference between Video-MCIBI++ and Video-MCIBI \cite{jin2021memory} is the adopted memory module. 
Specifically, the memory module in Video-MCIBI++ stores the dataset-level distribution information of each category while Video-MCIBI stores the dataset-level category representations. 
It is observed that UperNet+MCIBI++ with ResNet-101 achieves $1.10\%$ mIoU, $1.91\%$ TC and $1.51\%$ mVC$_{8}$ improvements than UperNet+MCIBI with ResNet-101,
and UperNet+MCIBI++ with the transformer-based backbone (Swin-Large) have the consistent improvements compared to UperNet+MCIBI.
These results are consistent with the previous observation in Section \ref{ImageSegmentorAblationAnalysis}, \ie, soft mining contextual information beyond image (MCIBI++) can generate more discriminative dataset-level category representations than mining contextual information beyond image (MCIBI) \cite{jin2021memory}.

\subsection{Comparison with the State-of-the-Art Methods}
Table \ref{sotavspw} also compares MCIBI++ and Video-MCIBI++ with the existing image or video semantic segmentation frameworks on the validation set of VSPW.
When adopting the ResNet-101 as the backbone, UperNet+MCIBI++ achieves $43.21\%$ mIoU, which outperforms the previous best segmentation framework TCB$_{\emph{st-ocr~mem}}$ by $5.39\%$ mIoU.
After introducing the temporal information and a more robust backbone network, \ie, Swin-Large, UperNet+Video-MCIBI++ reports the new state-of-the-art results, \emph{i.e.}, $58.88\%$ mIoU, $81.21\%$ TC and $91.53\%$ mVC$_{8}$.

\subsection{Quantitative Results}
Some segmentation examples from the VSPW validation set are demonstrated in Fig.~\ref{compare_vspw}.
We can observe that the results from Video-MCIBI++ (\emph{i.e.}, the third row) are more stable and accurate than MCIBI++ (\emph{i.e.}, the second row).
For example, in the first video (\emph{i.e.}, the first row), MCIBI++ labels one part of the bed as the other categories while Video-MCIBI++ can predict the bed accurately and stably.

\section{Application Discussion}
In this section, we discuss some applications of our proposed MCIBI++ paradigm and some potential issues we should pay attention to.
The first point is the potential domain divergence problem of different datasets. 
For instance, if the trained dataset only contains the single-person scenes, the dataset-level distribution information of each category in the memory module may be insufficient for the multi-person scenes.
Although the results in Table~\ref{generalization} show that the memory module can carry some high-level information to benefit the human parsing in different domains,
the overall segmentation performance of the segmentors trained on the single-person scenes is still far lower than the segmentors which are trained on the multi-person scenes directly.
Therefore, it is still necessary to design other strategies to tackle the domain divergence.
For example, during the network training, we suggest that the users can conduct some data augmentation~\cite{zou2022learning} to generate the training images of various domains so that the memory module can better adopt to the realistic scene.
Moreover, we find that some recent works~\cite{kim2022pin} have followed our previous study (\ie, the conference version) and designed some strategies to store the domain-agnostic information into the memory to achieve domain adaptation, which may be also very helpful for the application of our method. 
The second point we should pay attention to is that introducing memory module may make the network vulnerable from the adversarial attacks.
To alleviate the problem, we suggest that the users can also apply some data augmentation strategies \cite{zantedeschi2017efficient} during training to enhance the generation ability of the memory module.
Overall, we expect that proposed MCIBI++ can shine in the real applications.

\section{Conclusion and Outlook}
This paper presents an new perspective for tackling the context aggregation in semantic segmentation.
We propose to mine the contextual information beyond the input image in a soft way (MCIBI++) to further improve the pixel representations.
A memory module is set to store the dataset-level distribution information of various categories and 
a dataset-level context aggregation scheme is designed to aggregate the dataset-level category representations yielded by the stored distribution information for each pixel according to their class probability distribution.
At last, the aggregated dataset-level contextual information is leveraged to augment the original pixel representations for more accurate prediction.
In addition, a coarse-to-fine iterative inference strategy is proposed to further boost the segmentation performance by progressively improving the accuracy of aggregate weights.
Also, MCIBI++ is generalized to the video semantic segmentation task by incorporating the temporal information in the process of gathering the dataset-level category representations for the pixels.
Extensive experiments demonstrate the effectiveness of our method.

Comparing Table \ref{upperbound} and Table \ref{ablation}, we can observe that the achieved segmentation performance is far below the theoretical performance limit of our framework ($43.74\%$ \emph{v.s.} $68.82\%$).
We thus believe that our future works can focus on how to aggregate more accurate dataset-level category representations for the pixels, \ie, how to improve the dataset-level context aggregation scheme.
In addition, modeling the dataset-level category representations better and updating the memory module more efficiently will also be further investigated in our future work.



\section*{Acknowledgments}
The work described in this paper is supported in part by the HK RGC under T45-401/22-N and in part by HKU Seed Fund for Basic Research (Project No. 202009185079 and 202111159073).


\ifCLASSOPTIONcaptionsoff
  \newpage
\fi



\bibliographystyle{IEEEtran}
\bibliography{jrnl.bib}
\end{document}